%% file: collas2023_conference.tex
\documentclass{article} % For LaTeX2e
\usepackage{collas2023_conference,times}
\usepackage{easyReview}

\usepackage{algorithm}
\usepackage{algpseudocode}
\usepackage{xcolor}
\usepackage{subcaption} % for subfigures
\usepackage{booktabs} % for table
\usepackage{graphicx}
\usepackage{wrapfig}
\usepackage{multirow}

\usepackage[T1]{fontenc} % cater for names with non-english characters (e.g. continualworld bibtex)

\input{math_commands.tex}

\usepackage{hyperref}
\hypersetup{
    colorlinks=true,
    linkcolor=red,
    filecolor=magenta,
    urlcolor=blue,
    citecolor=purple,
    pdftitle={Overleaf Example},
    pdfpagemode=FullScreen,
    }

\def\codeURL{\url{https://github.com/DMIU-ShELL/deeprl-shell}}

\title{Sharing Lifelong Reinforcement Learning Knowledge via Modulating Masks}

\author{Saptarshi Nath\textsuperscript{1}, Christos Peridis\textsuperscript{1}, Eseoghene Ben-Iwhiwhu\textsuperscript{1}, Xinran Liu,\textsuperscript{2} \\
\textbf{Shirin Dora\textsuperscript{1}, Cong Liu\textsuperscript{3}, Soheil Kolouri\textsuperscript{2}, Andrea Soltoggio\textsuperscript{1}}\\
\\
\textsuperscript{1} Department of Computer Science, Loughborough University, United Kingdom \\
\textsuperscript{2} Department of Electrical Engineering and Computer Science, Vanderbilt University, United States \\
\textsuperscript{3} Department of Computer Science and Engineering, University of California, Riverside, United States \\
\\
\texttt{\{s.nath, c.peridis, e.ben-iwhiwhu, s.dora, a.soltoggio\}@lboro.ac.uk}, \\
\texttt{\{xinran.liu, soheil.kolouri\}@vanderbilt.edu}, \\
\texttt{congl@ucr.edu}
}

\preprintcopy % Uncomment for the preprint version, but NOT for submission.

\begin{document}

\maketitle

\begin{abstract}
Lifelong learning agents aim to learn multiple tasks sequentially over a lifetime. This involves the ability to exploit previous knowledge when learning new tasks and to avoid forgetting. Recently, modulating masks, a specific type of parameter isolation approach, have shown promise in both supervised and reinforcement learning. While lifelong learning algorithms have been investigated mainly within a single-agent approach, a question remains on how multiple agents can share lifelong learning knowledge with each other. We show that the parameter isolation mechanism used by modulating masks is particularly suitable for exchanging knowledge among agents in a distributed and decentralized system of lifelong learners. The key idea is that isolating specific task knowledge to specific masks allows agents to transfer only specific knowledge on-demand, resulting in a robust and effective collective of agents.  We assume fully distributed and asynchronous scenarios with dynamic agent numbers and connectivity. An on-demand communication protocol ensures agents query their peers for specific masks to be transferred and integrated into their policies when facing each task. Experiments indicate that on-demand mask communication is an effective way to implement distributed and decentralized lifelong reinforcement learning, and provides a lifelong learning benefit with respect to distributed RL baselines such as DD-PPO, IMPALA, and PPO+EWC. The system is particularly robust to connection drops and demonstrates rapid learning due to knowledge exchange. 
\end{abstract}
 
\section{Introduction}
\vspace{-5pt}

Distributed lifelong reinforcement learning (DLRL) is an emerging research field that offers scalability, robustness, and speed in real-world scenarios where multiple agents continuously learn. A key aspect of DLRL is the ability of multiple agents to learn multiple tasks sequentially while cooperating through information sharing. Notable advances have been made in related areas such as distributed reinforcement learning \citep{espeholt2018impala,wijmans2019dd}, federated reinforcement learning \citep{Qi2021}, and lifelong reinforcement learning \citep{zhan2017scalable,abel2018state,abel2018policy,xie2020deep}. While the integration of lifelong learning and distributed learning is emerging only recently \citep{mohammadi2019collaborative,songFLLEdge2022}, the combination of these two paradigms could yield efficient, scalable, and robust learning systems.

The use of multiple workers in RL has been applied to accelerate data collection in methods like A3C \citep{mnih2016asynchronous} while training a single model. Distributed reinforcement learning (DRL) \citep{weiss1995distributed,espeholt2018impala,wijmans2019dd} expands on the concept of multiple workers to include multiple agents and may utilize a fully decentralized approach where each agent learns a potentially different model. Such systems can exhibit high learning performance and robustness. However, learning across multiple agents does not address the issue of non-IID data, which frequently arises when multiple tasks are learned sequentially over time. 

Lifelong or continual learning specifically addresses the issue of sequential task learning, in which an agent learns multiple tasks characterized by unique input, transition, and reward distributions \citep{thrun1995lifelong,de2021continual}. The primary challenge is integrating a new task into existing knowledge without experiencing catastrophic forgetting \citep{mccloskey1989catastrophic}. Various approaches have been suggested to implement lifelong learning: \cite{de2021continual} propose a taxonomy with three main approaches, namely replay methods, regularization methods, and parameter isolation methods. Parameter isolation methods, in particular, allocate different parameters in a model to different tasks. In this paper, we suggest that such isolation mechanisms can be advantageous in a distributed setting where agents need to share knowledge.

We adopt a lifelong learning approach based on modulating masks, which have demonstrated competitive performance in supervised learning \citep{mallya2018packnet,zhou2019deconstructing,wortsman2020supermasks,koster2022signing_sm}. Recently, modulating masks have also been extended to lifelong reinforcement learning (LRL), enabling the integration and exploitation of previous knowledge when learning new tasks \citep{ben2022lifelong}. A natural question arising from these recent studies is whether decomposing knowledge into masks can be utilized to transfer task-specific information across agents, implementing a distributed system of lifelong reinforcement learners. In this paper, we propose a system where agents use a fully asynchronous and decentralized protocol to query each other and exchange only relevant information for their respective tasks. The proposed proof-of-concept, named \emph{lifelong learning distributed decentralized collective} (L2D2-C), suggests a lifelong learning approach to distributed decentralized learning. We test the proposed approach on benchmarks designed for multiple sequential tasks, namely the CT-graph \citep{soltoggio2023configurable} and Minigrid  \citep{gym_minigrid}. The simulations show that the system accelerates learning with respect to a single lifelong learning agent by a factor close to the number of agents. In addition, we demonstrate the system's robustness to connection drops. Comparisons with existing distributed RL approaches, namely DD-PPO \citep{wijmans2019dd} and IMPALA \citep{espeholt2018impala}, and non-distributed LL approaches (PPO+EWC) illustrate the advantages of integrating both LL and sharing. This study only illustrates initial experiments to suggest the validity of the idea, encouraging further future investigations for particular aspects of the system. The code to launch L2D2-C and reproduce the results is freely available at \codeURL.

The paper is structured as follows. Section \ref{sec:related_work} provides a review of related work and introduces the necessary background concepts. Section \ref{sec:method} introduces the \emph{lifelong learning distributed decentralized collective} (L2D2-C) approach. In Section \ref{sec:experiments}, we present the results of our simulations and analyze them. Finally, in Section \ref{sec:discussion}, we discuss the implications of our findings and conclude the paper. Additional information is provided in the Appendix.

%%%%%%%%%%%%%%%%%%%%%%%%%%%%%%%%%%%%%%
\section{Related Works and Background}
\label{sec:related_work}
\vspace{-5pt}

We focus on a relatively unexplored scenario in which multiple lifelong learning agents learn from their own non-IID streams of tasks and share relevant acquired knowledge with each other upon request. While individual aspects of such a system have been studied in the literature, our work brings together several of these concepts to create a more comprehensive understanding. For example, lifelong reinforcement learning (Section \ref{sec:LRL}) explores continual learning methods in RL settings for a single agent, while federated learning (Section \ref{sec:FL}) investigates the use of centralized and decentralized methods to learn with non-IID data. Finally, distributed RL (Section \ref{sec:DRL}) seeks to enhance the speed of RL by deploying multiple agents working together. In addition, the concept of modulating masks has become crucial in various fields, including lifelong learning. In Section \ref{sec:maskRL}, we examine how modulating masks can be effectively employed to facilitate lifelong reinforcement learning.

\subsection{Lifelong Reinforcement Learning}
\vspace{-3pt}
\label{sec:LRL}

Lifelong learning, a concept known for decades \citep{thrun1995lifelong}, has gained prominence as an active field of study in recent neural network studies \citep{soltoggio2018born,van2019three,hadsell2020embracing, khetarpal2020towards, de2021continual}. The desiderata of a lifelong learner lie in its ability to learn multiple tasks in sequence while avoiding catastrophic forgetting \citep{mccloskey1989catastrophic}, enabling knowledge reuse via forward and backward transfer \citep{chaudhry2018riemannian} and reducing task interference \citep{kessler2022state}. %, and enabling efficient use of the model capacity.

Different approaches can be largely categorized into regularization methods, modular methods, or a combination of both. Regularization methods apply an additional penalty term in the loss function to penalize either large structural changes or large functional changes in the network. Structural regularization or synaptic consolidation methods \citep{kirkpatrick2017ewc, zenke2017si, aljundi2018mas, kolouri2019scp} penalizes large change weights in the network that are important for maintaining the performance of previously learned tasks. Modular methods learn modules or a combination of modules useful for solving each task. A module is expressed as either: (i) a mask that activates sub-regions of a fixed neural network \citep{mallya2018packnet, serra2018hat, mallya2018piggyback, wortsman2020supermasks, koster2022signing_sm, ben2022lifelong}, or (ii) a neural network that is expanded \citep{rusu2016progressive, yoon2018den} or compositionally combined \citep{mendez2021lifelong, mendez2022neuralcomposition} with other modules as new tasks are learned.

Lifelong RL agents are set up as a combination of standard deep RL and lifelong learning algorithms. In \cite{kirkpatrick2017ewc}, EWC was combined with DQN \citep{mnih2013playing} to solve Atari games \citep{bellemare2013atari}. Progress \& Compress \citep{schwarz2018pnc} showed further extensions of EWC by combining EWC with IMPALA \citep{espeholt2018impala} and policy distillation techniques.  In CLEAR \citep{rolnick2019clear}, IMPALA was combined with a replay method and behavioral cloning. SAC \citep{haarnoja2018sac} was combined with a number of lifelong learning algorithms in \cite{wolczyk2021continualworld}. For offline RL, \cite{xie2022lifelong} leveraged a replay buffer in SAC and importance weighting technique to specifically tackle forward transfer in a robotics agent, while \cite{mendez2022neuralcomposition} merged neural modular composition, offline RL, and PPO \citep{schulman2017ppo} to enable reuse of knowledge and fast task learning. For masking approaches, \cite{wolczyk2021continualworld} combined SAC and PackNet \citep{mallya2018packnet} (masks derived from iterative pruning), while \cite{ben2022lifelong} demonstrated the use of directly learned modulating masks with PPO and introduced a knowledge reuse mechanism for the masking agents.

\subsection{Federated Learning for non-IID tasks}
\vspace{-3pt}
\label{sec:FL}

Federated learning (FL) generally considers multi-modal systems that train centralized models across various nodes, enabling accelerated training on massive amounts of data \citep{kairouz2021advances}. Data anonymity and security are one of the core concerns in federated learning. FL approaches that address fully decentralized and non-IID scenarios are focused primarily on reducing the efficiency drops and guaranteeing convergence for a model for a single task. Sources of non-IID data in FL are often related to the geographical distributions of the clients, time of training, and other factors that relate often to biasing aspects of the data collection rather than the assumption of different tasks. As a result, rather than employing lifelong learning algorithms, more often FL on non-IID data adopts solutions to mitigate the effect of different distributions such as sharing sample data sets across clients \citep{wang2018dataset,zhao2018federated,pmlr-v199-ma22a}. An alternative FL approach to manage non-IID data is to assume the presence of multiple tasks: in such cases, FL integrates aspects of multi-task learning and meta-learning. In multi-task learning, the result of the process is one model per task \citep{NIPS2017_6211080f,zhang2022federated}. Finally, meta-learning \citep{hospedales2021meta} can be effective to learn a global model for further fine-tuning to local data. Combinations of meta-learning approaches such as MAML \citep{finn2017model} with FL have been explored in \cite{jiang2019improving,khodak2019adaptive}. The objective of such approaches is often that of adapting a global model to a specific local data set \citep{fallah2020personalized,li2021ditto,tan2022towards} in what is referred to as personalized FL.

%%%%%%%%%%%%%%%%%%%%%%%%%%%%%%%%%%%%%%%%%%%%%%%
\subsection{Distributed Reinforcement Learning}
\vspace{-3pt}
\label{sec:DRL}

The concept of distributed reinforcement learning (DRL) was initially introduced to distribute computation across many nodes and thereby increase the speed at which data collection and training are performed \citep{weiss1995distributed,gronauer2022multi}. Both synchronous and asynchronous methods such as A2C/A3C \citep{mnih2016asynchronous}, DPPO \citep{heess2017emergence}, IMPALA \citep{espeholt2018impala}, and  others, use multiple workers to increase the rate of data collection while training a central model. A variation of PPO \citep{schulman2017ppo}, DD-PPO \citep{wijmans2019dd} extends the framework to be distributed and uses a decentralized optimizer, reporting a performance that scales nearly linearly with the number of nodes. While these algorithms have shown improvements in the SoTA on various benchmarks, they do not incorporate lifelong learning capabilities. %\add{The incorporation of lifelong learning algorithm in distributed RL is emerging in the last few years \citep{papoudakis2019dealing}}.

%%%%%%%%%%%%%%%%%%%%%%%%%%%%%%%%%%%%%%%%%%%%%%%%%%%%%%%%%%%%%%%%%
\subsection{Modulating Masks for Lifelong Reinforcement Learning}\vspace{-3pt}
\label{sec:maskRL}

The idea of using modulation in RL tasks is not new \citep{doya2002metalearn_neuromod,dayan2008reinforcement,soltoggio2008evolutionary,ben2022context}, but recently developed masking methods for deep supervised lifelong learning, e.g., \cite{wortsman2020supermasks}, have shown the advantages of isolation parameter methods. In \cite{ben2022lifelong}, modulating masks are shown to work effectively in deep reinforcement learning when combined with RL algorithms such as PPO \citep{schulman2017ppo} or IMPALA \citep{espeholt2018impala}. Other approaches such as \cite{mallya2018packnet} or \cite{mallya2018piggyback} use masks in RL, but modify the backbone network as well, making them less suitable for knowledge sharing. In \cite{ben2022lifelong} the agent contains a neural network policy $\pi_{\theta, \Phi}$, parameterized by the weights (backbone) of the network $\theta$ and the mask score parameters $\Phi=\{\phi_1, \ldots \phi_k\}$ for tasks $1 \ldots k$. The backbone is randomly initialized (using the signed Kaiming constant \citep{ramanujan2020s}) but kept fixed and remains unchanged, while mask score parameters are optimized during learning and applied on the backbone. For any given task $k$, $\phi_k$ represents the mask parameters across all layers of the network. In a network consisting of $L$ layers, $\phi_k = \{S_k^1, \ldots S_k^L\}$ where $S_k^i$ is the mask parameter in layer $i$ for task $k$. To apply a mask on the backbone, $\phi_k$ is quantized using an element-wise threshold function (i.e., 1 if $\phi_{k, \{i, j\}} > 0$, otherwise 0) to generate a binary mask that is element-wise multiplied with $\theta$. The multiplicative process activates and deactivates different regions of the backbone. Also, the framework supports knowledge reuse via the weighted linear combination of previously learned task masks and the current task mask. For each layer $l$, this is expressed as: 
\begin{equation}
    S^{l, lc} = \left( \sum_{i=1}^{k}\beta_i^{l} S_{i}^{l, *}\right ) + \beta_{k+1}^{l}S_{k+1}^{l}\quad,
    \label{eq:mask}
\end{equation}
where $S^{l, lc}$ denotes the transformed mask score parameters for task $k+1$ after the linear combination step. $S_i^{l, *}$ denotes the optimal mask score parameters for previously learned task $i$, and $\beta_1^{l}, \ldots \beta_{k+1}^{l}$ are the linear coefficients (weights) of the operation at layer $l$. $\mathrm{Softmax}$ is applied on the linear co-efficient to normalize it and also express the degree to which each mask is relevant to learn the current task. The learned parameters in the framework are $\beta$ and $\Phi$. Once the mask for task $k + 1 $ has been learned using linear combination parameters, it is consolidated into a mask by itself, and thus the masks of other tasks can undergo further changes. %For completeness, the algorithm of the modulating mask LRL is presented in Appendix \ref{apndx:mask-lrl-algo}. 
The approach has been shown to be an effective way to implement LL by avoiding forgetting via complete parameter isolation and exploiting previous knowledge to facilitate learning of future tasks in a sequential curriculum. 

%%%%%%%%%%%%%%%%%%%%%%%%%%%%%%%%%%%%%%%%%%%%%%%%%%%%%%%
\section{Sharing masks across lifelong learning agents}\vspace{-5pt}
\label{sec:method}

The main research questions we want to address are, can we effectively transfer modulating masks across agents to produce a distributed lifelong learning system? Are masks a suitable form of parameter isolation that allows for task-specific policies to be transferred and integrated across agents? If we do so, can we observe a learning benefit from the synergy of both lifelong learning and sharing? The benefits may include increased learning speed of the collective with respect to the single agent, albeit at the increased overall computational resources. A hypothesis is that, if learning and sharing is efficient, $n$ agents may be up to $n$-times faster at learning a given curriculum than the single agent. We provide further technical considerations on such an idea of performance improvement in Section \ref{apndx:performance}. To address the objectives above, we designed an algorithm for communication and sharing that can transfer the relevant modulating masks from the appropriate agents when needed. To produce a fully asynchronous and fully distributed system, each agent needs to act without central coordination and independently determine what information to ask for and when. %In the following sections, we introduced the lifelong learning distributed decentralized collective (L2D2-C).  

\subsection{What and when to share}\vspace{-3pt}
\label{sec:whatwhen}

We assume that each agent in the system shares the same backbone network $\theta$ that is randomly initialized. Each agent stores a list of masks $\Phi=\{\phi_1, \ldots \phi_k\}$ for tasks that they have encountered, and therefore they can independently learn a subset of all available tasks. A research question is whether agents can opportunistically and timely exchange learned masks to (i) acquire the knowledge of a task that has been already learned by another agent and integrate it into its own knowledge (ii) continue to learn on that task in cooperation with other agents, thereby increasing the knowledge of the collective system. %Key to this idea is to devise a communication protocol that exploits the collective learning dynamics to create a fully decentralized and asynchronous system. 
With that aim, we introduce two main communication operations: queries and mask transfers. 

\textbf{Queries.} When Agent 1 faces a task, it sends a query to all other agents (IDQ) communicating the task representation (e.g., task ID) and its performance on that task. Agents that have seen that particular task before, and exceed the performance of Agent 1%by a threshold (e.g., at least 10\% better)
, send a query-response (QR) with their performance measures, effectively communicating that they have better knowledge to solve that particular task. All other agents do not answer the query.

\textbf{Mask transfers.} Agent 1 then selects the best agent that reports the highest performance on the task and requests the mask with mask-request (MR) sent only to that particular agent, which, in turn, answers by sending the requested mask with a mask-transfer (MTR). In the present implementation, we transfer the scores of Eq.\ \ref{eq:mask}. Transferring the sparse and binarized masks offers significant bandwidth saving, but requires a re-initialization of the scores to resume training. Note that the scores are not used in forward passes, so such a process does not cause a drop in performance. The optimization of communication, including different degrees of sparsity of masks could be the topic of future application-specific studies.

\begin{figure}
    \centering
        \includegraphics[width=0.9\textwidth]{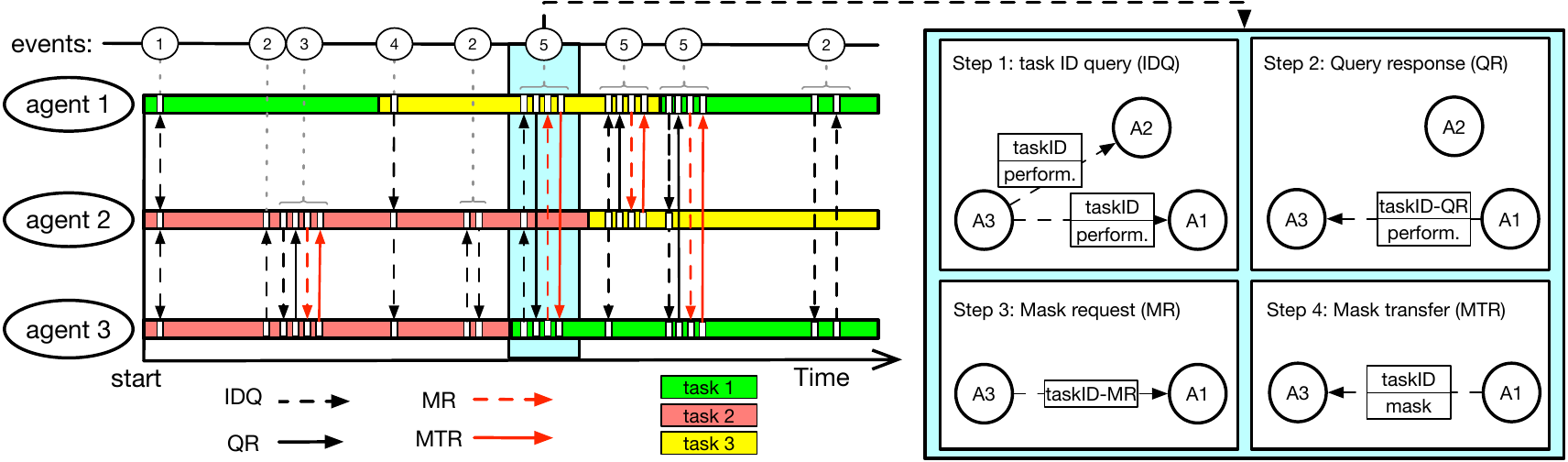} %&  %\includegraphics[width=0.58\textwidth]{ShELL Paper/figures/agentComm.pdf} \\
    \caption{Messages and steps in the communication protocol. (Left) Event 1: All agents send IDQ to all other agents. Event 2: Agent 3 sends an IDQ to agent 2, there is no answer because agent 2 cannot exceed agent 3 performance. Event 3:  An IDQ triggers a QR, followed by an MR and MTR. Event 4: Agent 1 starts a new task and issues IDQs, but there are no answers as no one has seen this task before. Event 5: Agent 3 starts task 1, issues IDQs, and receives a QR, which triggers an MR and MTR. (Right) A detailed description of  event 4. An IDQ message is sent by agent 3 when starting to learn task 1. A QR is sent back by agent 1 who has seen task 1 before and exceeds agent 3 performance. Agent 3 then selects the best-performing policy and requests the corresponding mask to agent 1 with an MR message. In step 4, agent 1 sends the requested mask to agent 3 (MTR). }
    \label{fig:comm}
\end{figure}

In summary, when an IDQ is triggered, the following steps take place:
\begin{itemize}\setlength{\itemsep}{-1pt}
    \item[Step 1:] Agent 1 sends a query (IDQ) to all other agents with the task ID and its performance.
    \item[Step 2:] Agents that have encountered that task before and exceed the performance by at least 10\% (estimated as the return over the last 512\footnote{Assuming the product of the rollout length and number of workers in tables \ref{tab.baseline_hyperparameters_l2d2-c} and \ref{tab.hyperparameters} is 512.} steps), respond with a query-response (QR).
    \item[Step 3:] Agent 1 sorts all the responses to find the agent X that has the highest performance and requests a mask from that particular agent (via a mask-request (MR)).
    \item[Step 4:] Agent X sends the mask corresponding to that task ID back to agent 1 with a mask-transfer (MTR).
\end{itemize}
The process is graphically illustrated in Figure  \ref{fig:comm}.
An IDQ is triggered by one of two events: a task change or a maximum length of independent learning. In the first case (task change), an agent is about to start learning a task and therefore issues IDQs to consult the collective about whether a policy (mask) exists that solves the task better than its own mask. The second case, instead, is designed to take advantage of more frequent communication among agents learning the same task, which periodically check the collective for better policies than their own.

The number of IDQ messages sent across the network depends on the frequency $f$ of IDQ messages per agent times $n$ the number of agents, with a growth of $fn^2$. The growth is quadratic with the number of agents in a fully connected network. However, such messages are small as they contain only the task ID. Moreover, they are sent only to reachable online agents. The number of QR messages is a fraction of the IDQ messages that depends on how many agents exceed the performance on that task. These are also small messages containing only the task ID and a measure of performance. The number of mask requests and responses grows linearly with the frequency of IDQs and the number of agents, i.e., $fn$. This is particularly important for the transfer of a mask, which is the largest type of communication that transfers the model parameters of Eq.\ \ref{eq:mask}.

\subsection{Knowledge distribution, performance and evaluating agents}\vspace{-3pt}
\label{sec:knowledge}

In a distributed system, determining how the collective's knowledge is distributed across agents is not straightforward. The communication protocol devised in Section \ref{sec:whatwhen} implies that agents only maintain knowledge of the tasks that they have encountered, and update such knowledge from the collective only when they re-encounter that task. We devised the following system to assess the performance of the system as a whole.  Given a curriculum $\mathcal{T} = \{\tau_1, \tau_2, \ldots, \tau_n\}$, a measure of performance can be expressed as the degree to which an agent can solve those tasks after a given learning time $\mu$. In other words, the learning objective can be set to maximize the reward across all tasks, i.e., a performance value $p$ can be defined as: \begin{equation}
    p(\mu) = \sum_{\tau=1}^{\tau=n}\sum_{t=\mu}^{\mu+\kappa} r(s_t, a_t)\quad,
    \label{eq:p}
\end{equation}
where $\mu$ is the time at which the system is assessed, $\kappa$ is the duration of an evaluation block set as a multiple of an episode duration, and $r(s_t, a_t)$ is the reward collected at each time step. If we divide $p(\mu)$ by the number of episodes in an evaluation period $\kappa$ (nr.\ of episodes in an evaluation $ee$), we obtain an instant cumulative return (ICR) of an agent at time $\mu$, i.e., $\mathrm{ICR}(\mu) = p(\mu)/ee$.  Eq.\ \ref{eq:p}, when computed over time also provides a basic but effective lifelong learning metric. While many other factors can be considered to assess lifelong learning systems \citep{new2022lifelong,baker2023}, $p(\mu)$ in Eq.\ \ref{eq:p} reveals how well an agent can solve all tasks after a period of training. 

To assess the performance of the learning collective with respect to a single agent, we also introduce two comparison values: (1) a performance advantage (PA) defined as the ratio $\mathrm{PA} = p_c(\mu)/p_s(\mu)$, where $p_c$ refers to a collective of agents and $p_s$ refers to a single agent: this is the advantage of a collective of agents with respect to a single agent; (2) a time advantage (TA), defined as the ratio between time-to-performance-p for the collective with respect to the single agent. This is how much faster the collective reaches a determined performance. 

Eq.\ \ref{eq:p} requires the unrolling of the sequence of tasks $\mathcal{T}$ to at least one agent in the collective. A common approach to test lifelong learning \citep{baker2023} is to stop learning to perform evaluation blocks, during which the agent is assessed. In our case, stopping any agent in the collective from learning in order to assess the performance will affect the performance itself. 
The solution we devised is to create one or more special agents called \emph{evaluating agents (EA)} with the following characteristics: the EA does not learn and does not answer queries, so it is invisible to the collective and it cannot impact their performance. However, it can query all  agents and fetch their knowledge, which is then tested continuously. The use of EAs is a way to monitor the performance of the collective without interfering with its learning dynamics. In the experiments in which we used the ICR metric (derived by Eq.\ \ref{eq:p}), one evaluation agent was deployed.

\subsection{Decentralized dynamic network and agent architecture}\vspace{-3pt}

To ensure that L2D2-C is completely decentralized and asynchronous, the entire architecture is contained with one agent, i.e., there is no coordinating or central unit. Each agent in the system is identified by their unique tuple <IP,Port>. An agent stores the following data structures: \emph{entry-points} and \emph{online-agents}. \emph{Entry-points} is a list of <IP,Port> tuples that lists all known agents. The list is continuously updated by adding any new agent that makes contact via an IDQ. When an agent enters the collective and makes first contact with another agent, it receives the \emph{entry-points} list from that agent, thus acquiring knowledge of the identities of the other agents. If an agent goes offline for a period of time, when is back online, it will attempt to access the collective by contacting the agents in this list. The \emph{online-agents} list keeps track of agents that are currently online, and it is used to send out IDQs. This system allows agents to (i) dynamically enter and leave the collective at any time and (ii) manage sparsely connected or constrained topologies by communicating only with reachable (i.e. online) agents. Such a simple setup allows L2D2-C to run multiple agents on the same server, across multiple servers, and across multiple locations anywhere in the world. 

As L2D2-C emerges from instantiating multiple single agents, the entire system architecture is contained within one agent. In the most general case of DLRL, events such as task changes, data collection and learning, and communication among agents, are stochastic and independent events. Thus, an agent is required to perform different operations in parallel (data collection, learning, and communication as both client and server) to maximize performance. L2D2-C is implemented with multi-processing and multi-threading to ensure all such operations can run simultaneously.
An overview of the agent's architecture is provided in Figure \ref{fig:architecture} in the Appendix \ref{sec:HWSW}.

%%%%%%%%%%%%%%%%%%%%%
\section{Experiments}
\label{sec:experiments}
\vspace{-5pt}
We present the tests of L2D2-C when multiple agents learn randomized curricula of tasks. The following metrics are measured: performance of a varying number of agents with respect to the  single lifelong learning agent (Section \ref{sec:comparison}); lifelong learning performance of L2D2-C with respect to IMPALA, DD-PPO, PPO, PPO+EWC (single-head) and PPO+EWC (multi-head) (Section \ref{sec:baselines}); performance of L2D2-C with unreliable communication (Section \ref{sec:robustness}).

\subsection{Benchmarks}\vspace{-3pt}
\label{sec:benchmarks}
We employed two benchmarks for discrete RL scenarios that are particularly suitable for sequential LL, the configurable tree graph (CT-graph) \citep{soltoggio2023configurable} (code at \cite{soltoggio2019ctgraph}) and the Minigrid \citep{gym_minigrid} environments. The CT-graph is a generative environment that implements a tree graph with 2D images which makes it suitable for evaluating lifelong RL approaches. A number of tasks with varying complexity can be automatically generated to create a large variety of curricula. It features sparse rewards and arbitrarily long episodes in different configurations. Due to the varying reward location, the CT-graph implements interfering tasks. Graphical illustrations of the environment are provided in Appendix \ref{apndx:env-ctgraph}. The Minigrid \citep{gym_minigrid} is a grid-world navigation environment, consisting of a number of predefined partially observable tasks with varying levels of complexity. The agent is required to reach a defined location in the grid world, avoiding obstacles such as walls, lava, moving balls, etc. The experiment protocol employs a curriculum of three tasks that consists of the following: $\mathrm{SimpleCrossingS9N1}$, $\mathrm{SimpleCrossingS9N2}$, $\mathrm{SimpleCrossingS9N3}$. Screenshots of all tasks are reported in the Appendix (\ref{apndx:env-minigrid}).

\subsection{L2D2-C versus a single lifelong learner (L2D2-C with one agent)}\vspace{-3pt} \label{sec:comparison}
\begin{figure}[t]
    \centering
    \begin{tabular}{cccc}
    \multicolumn{2}{c}{
        \includegraphics[width=0.4\textwidth]{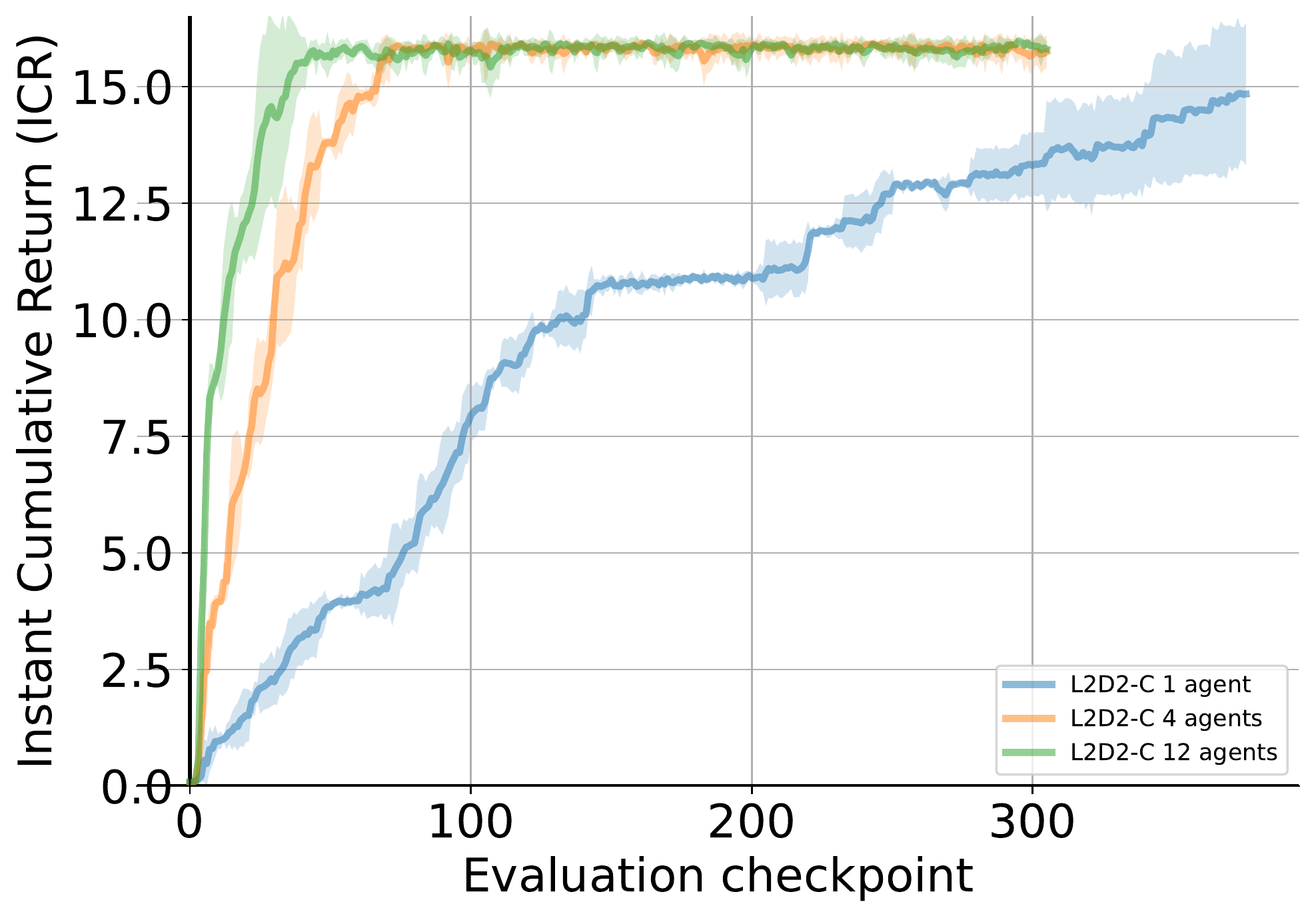}} &
        \multicolumn{2}{c}{
        \includegraphics[width=0.4\textwidth]{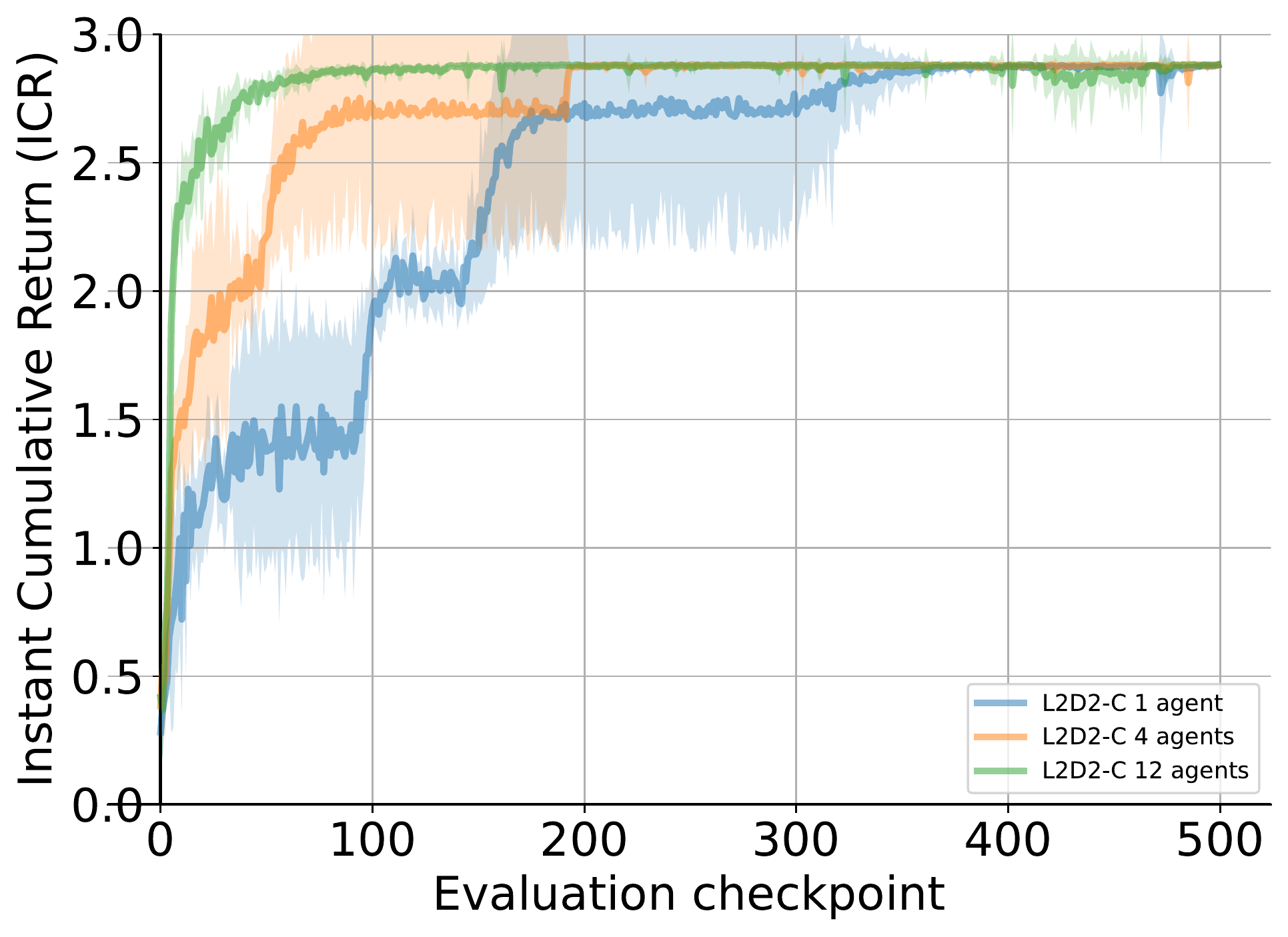}} 
       \\
        \multicolumn{2}{c}{(A)}& \multicolumn{2}{c}{(B)}\\              \includegraphics[width=0.22\textwidth]{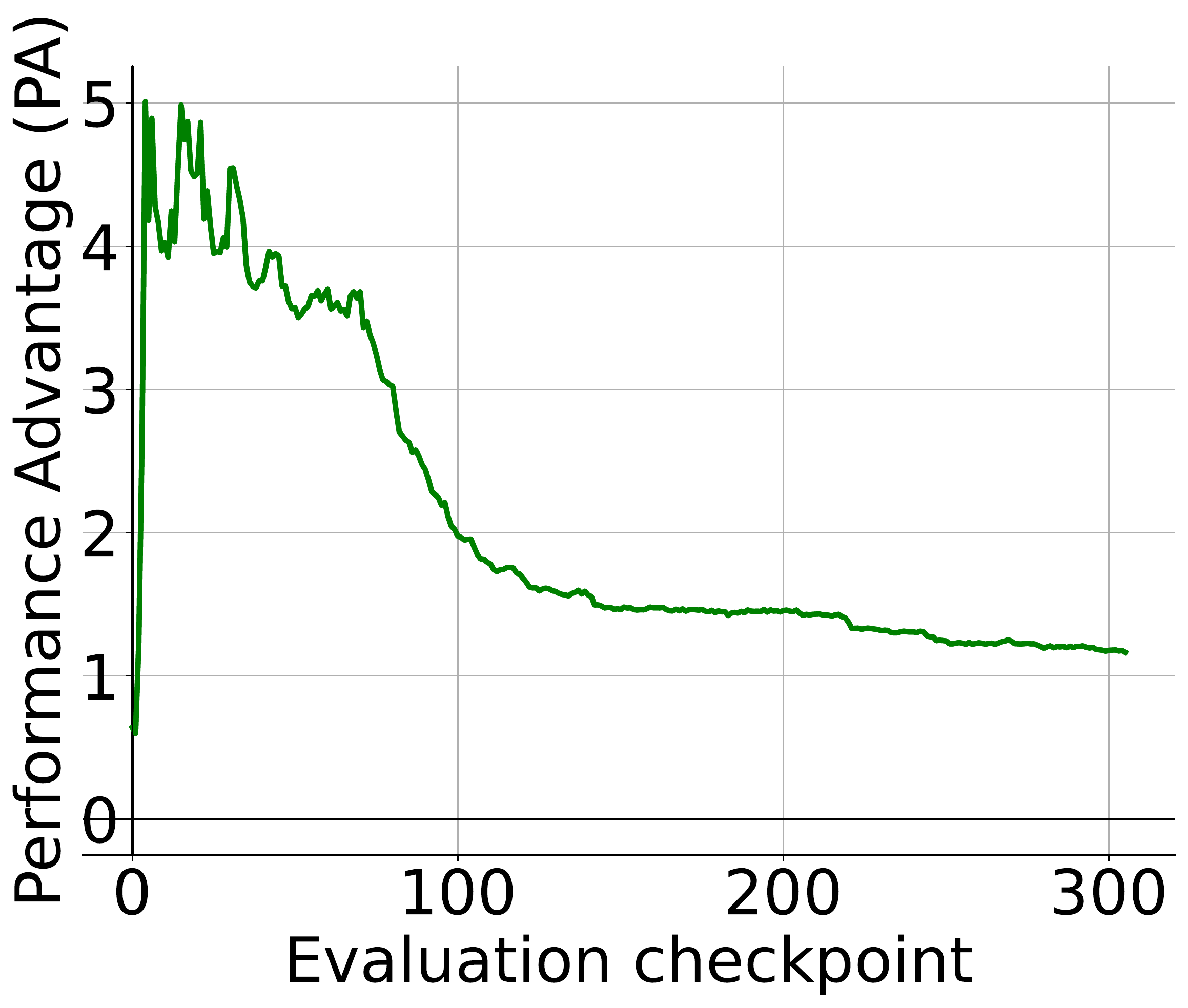} & 
        \includegraphics[width=0.22\textwidth]{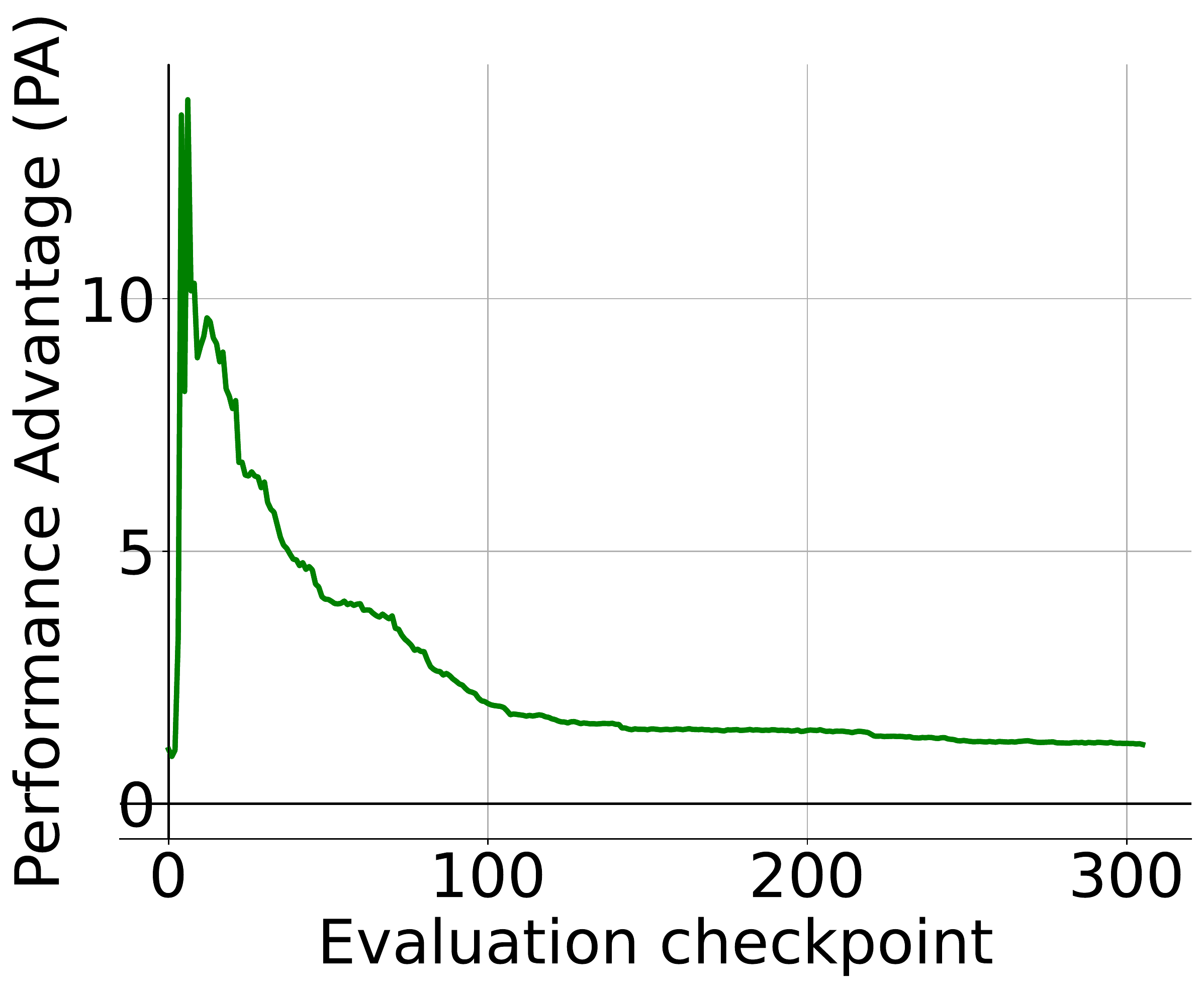} & \includegraphics[width=0.22\textwidth]{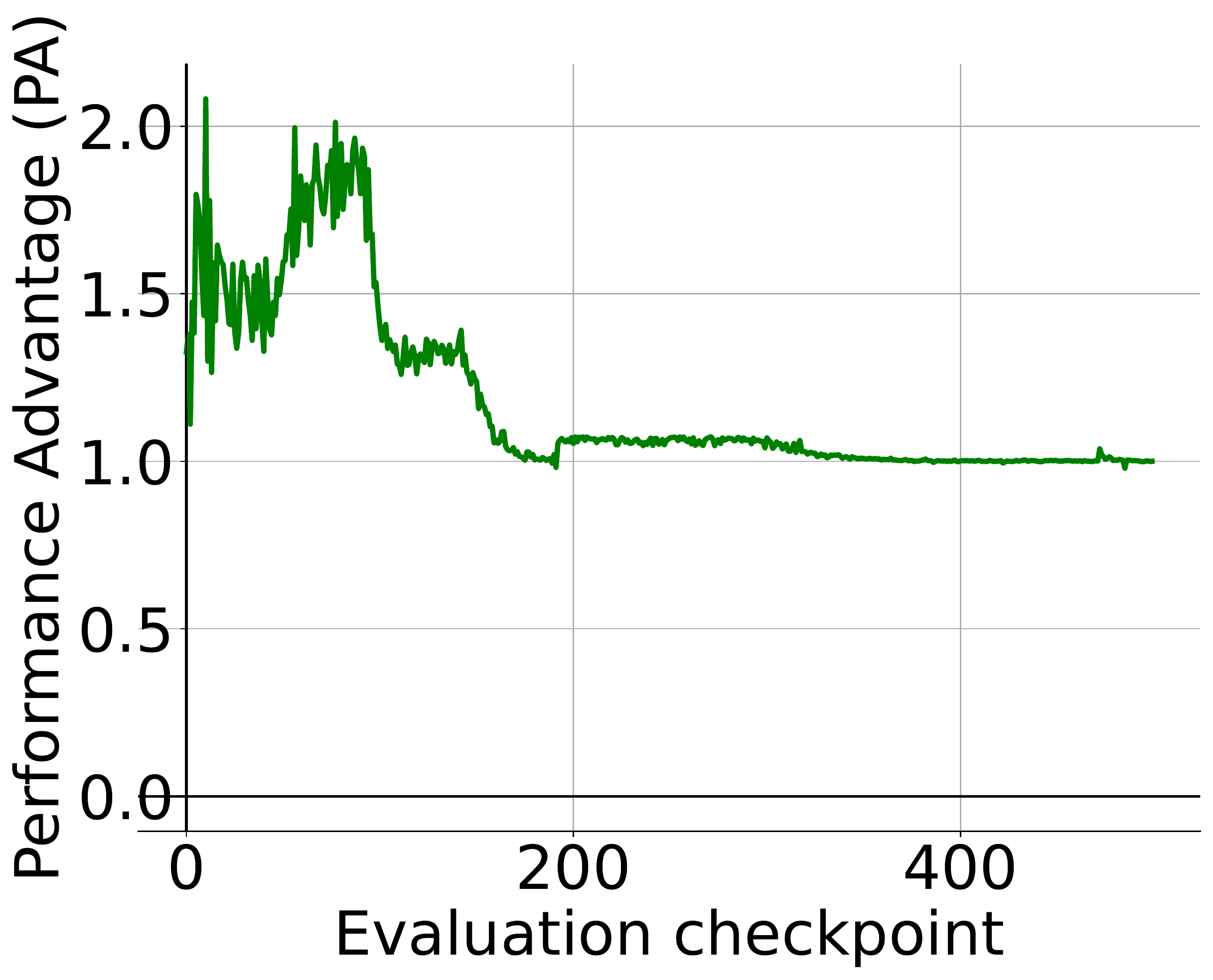} & \includegraphics[width=0.22\textwidth]{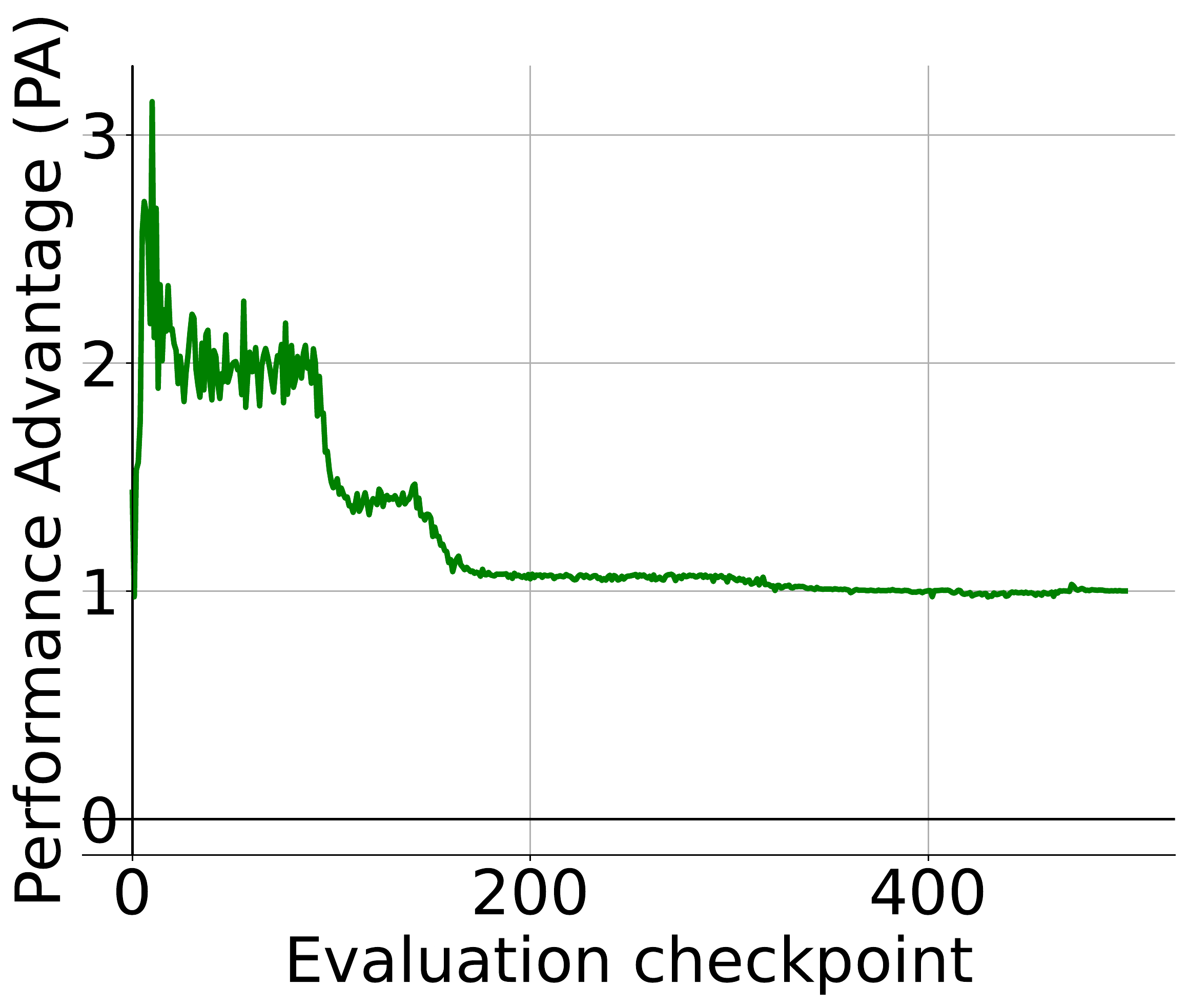}\\
       (C)&(D)&(E)&(F)\\
        \includegraphics[width=0.22\textwidth]{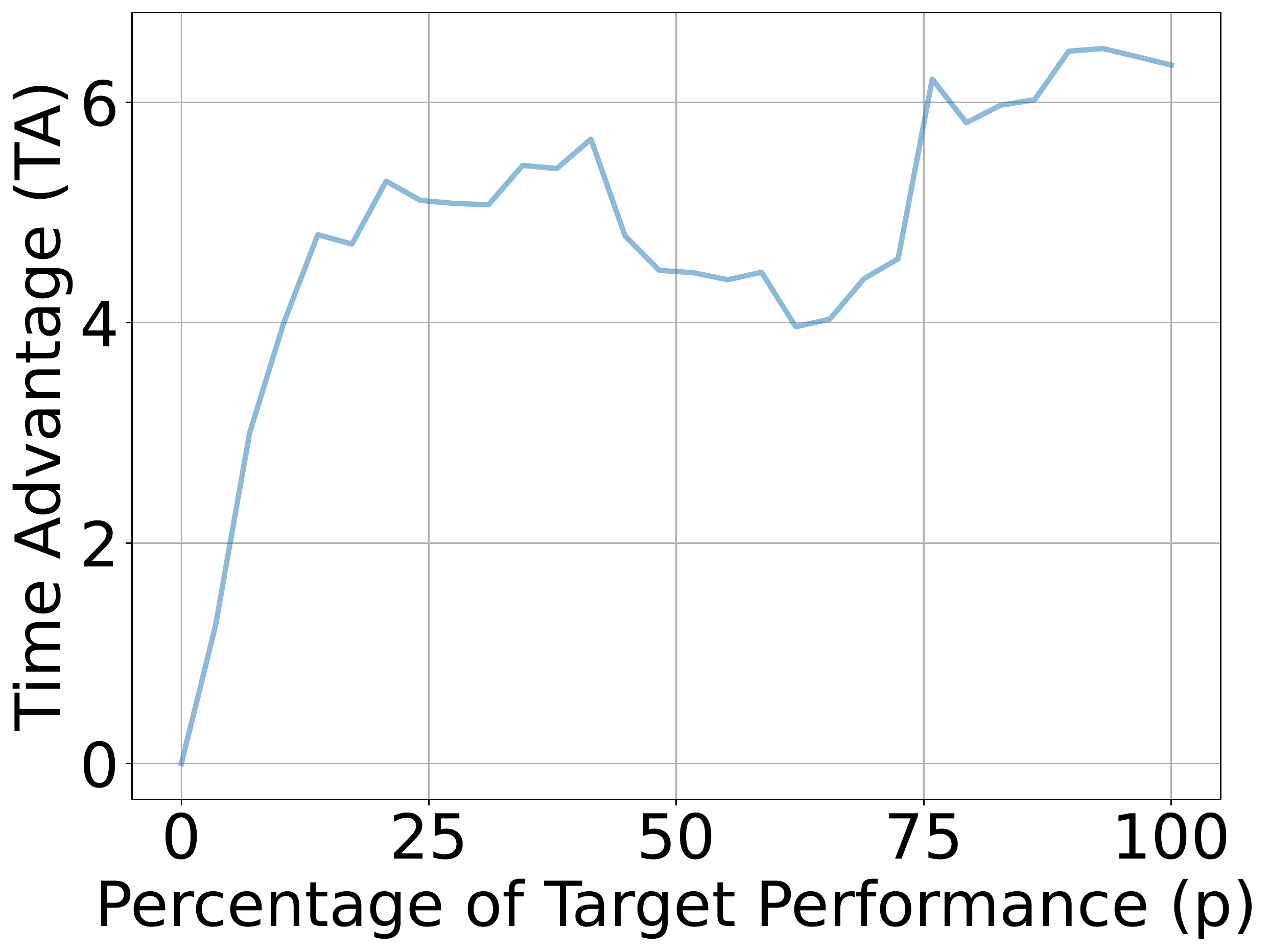} & 
        \includegraphics[width=0.22\textwidth]{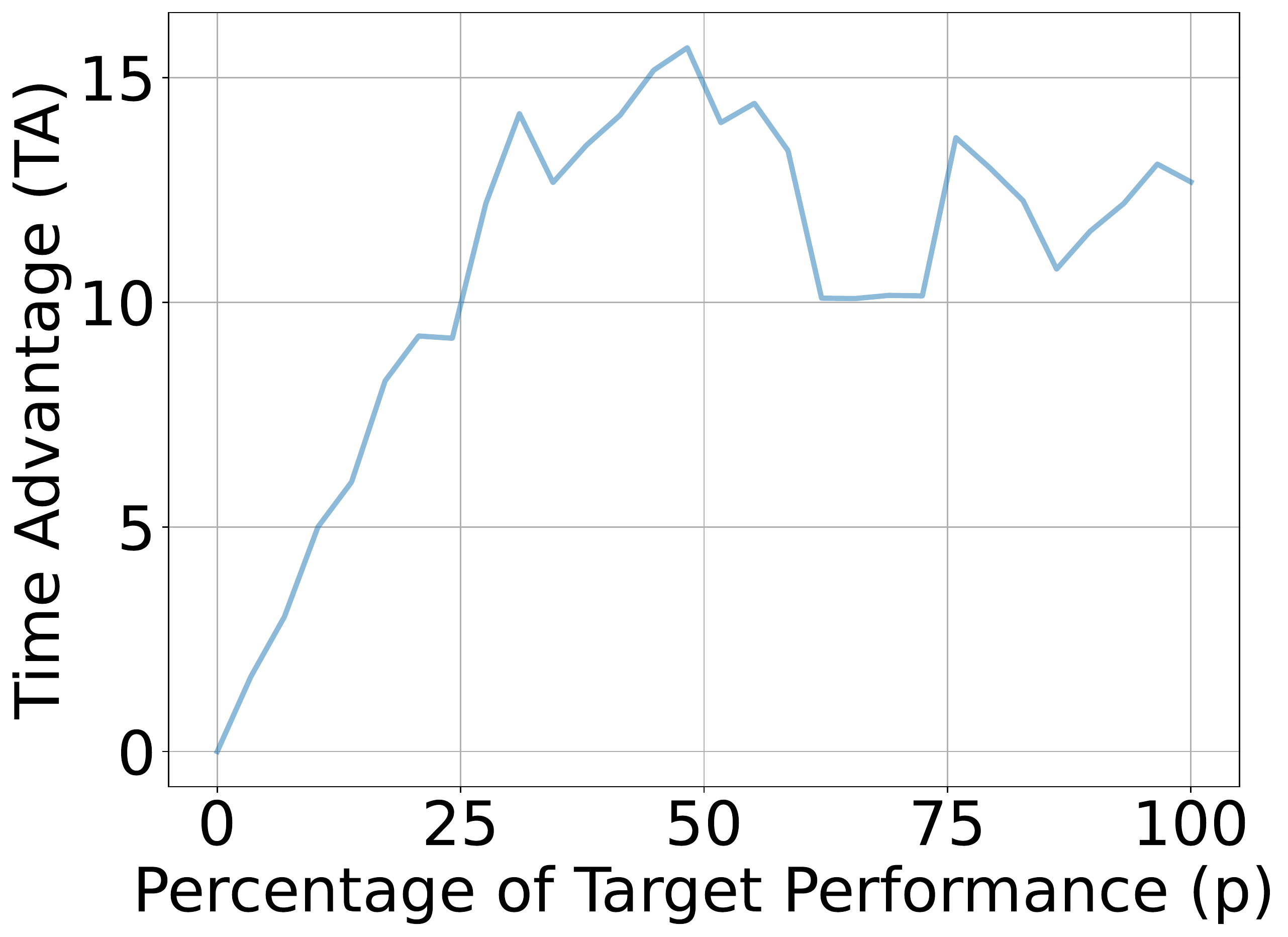} &
        \includegraphics[width=0.22\textwidth]{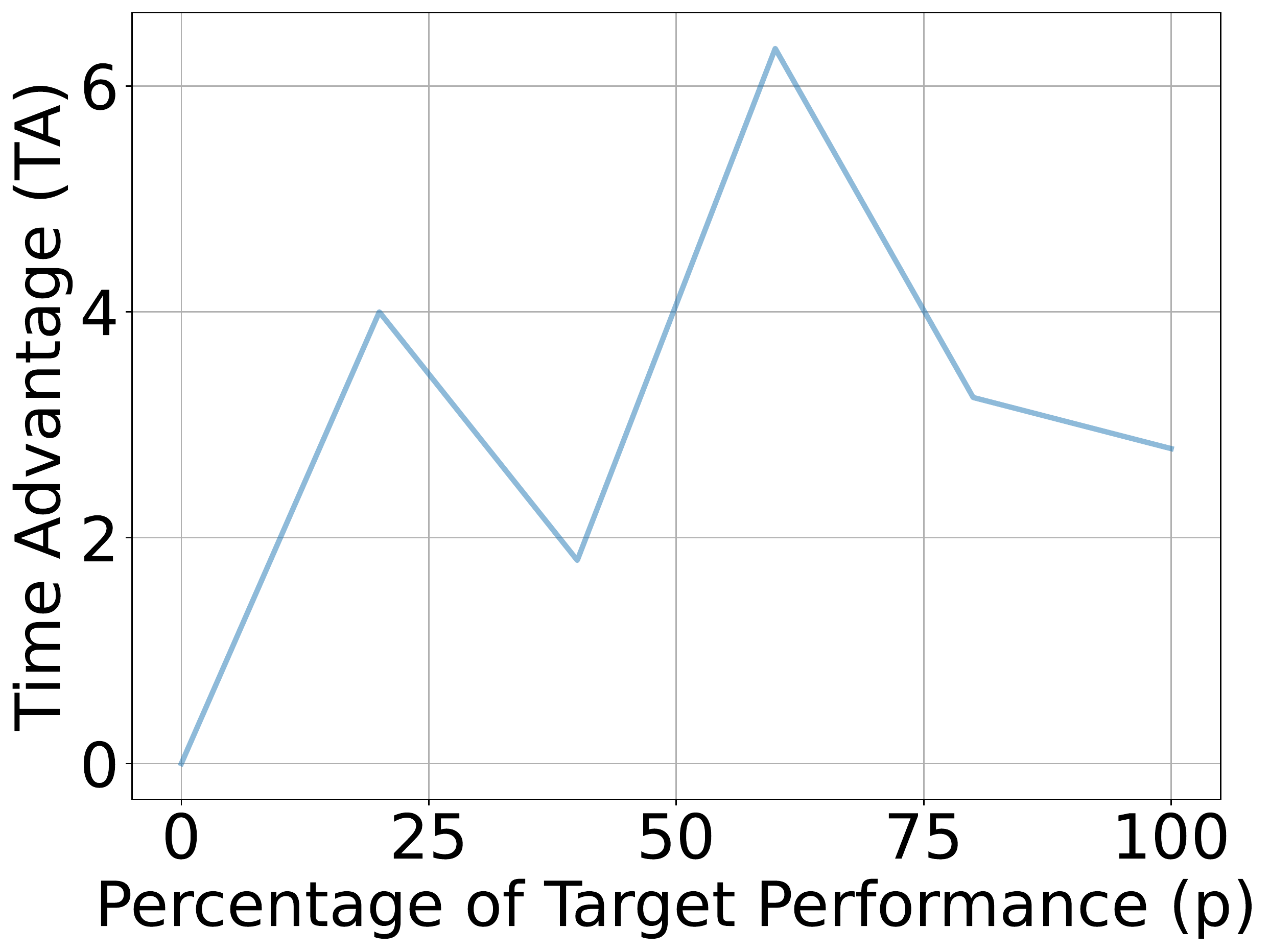} &
        \includegraphics[width=0.22\textwidth]{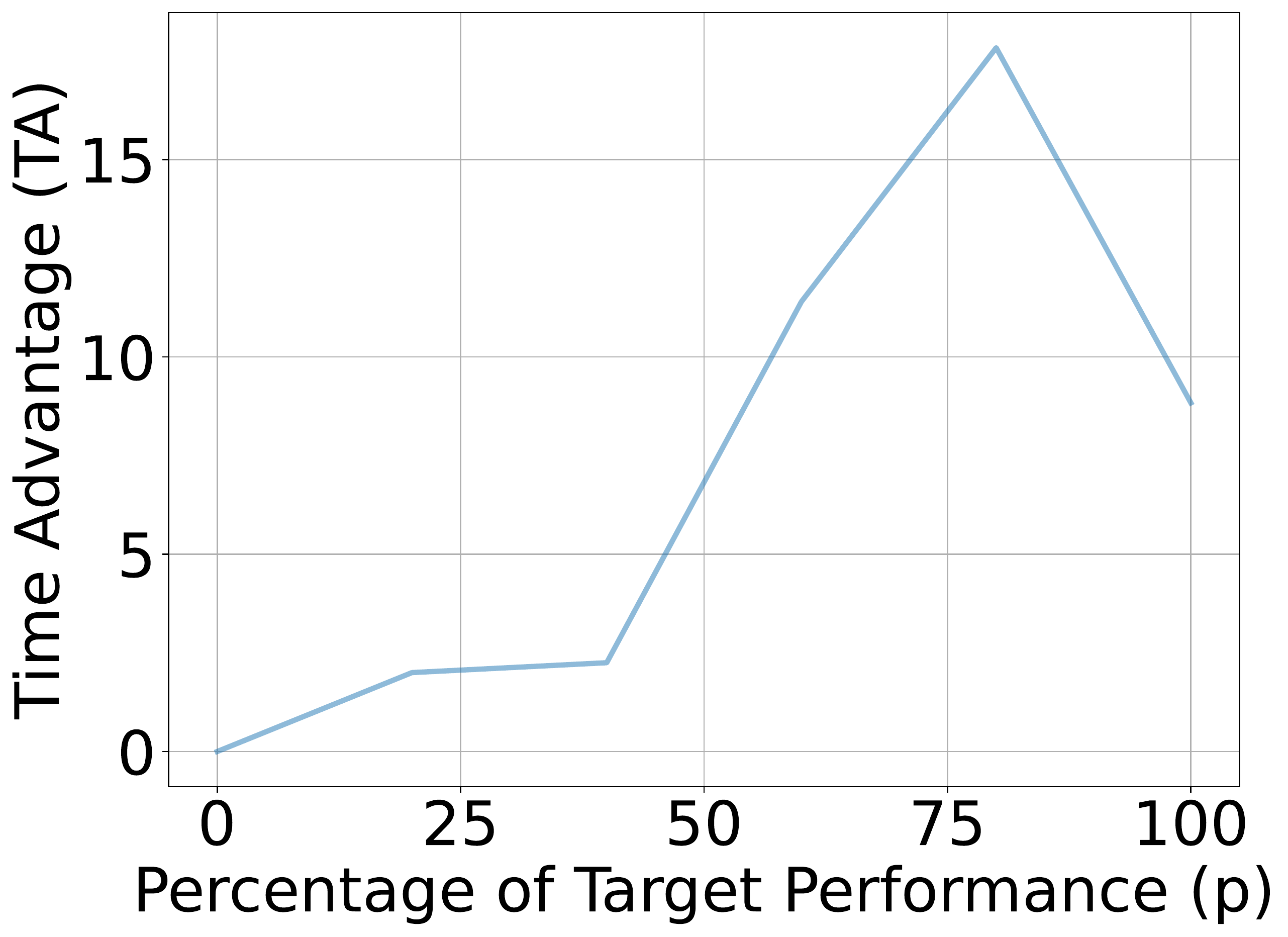} \\
        (G)&(H)&(I)&(J)\\
    \end{tabular}
    \caption{L2D2-C versus a single agent. (A) The instant cumulative return (ICR) provides the sum of the performance of the evaluating agents on 16 CT-graph tasks. A single agent is compared with a 4 and a 12-agent collective. (B) ICR on 3 Minigrid tasks. (C, D, E, F) Performance advantage (PA) of L2D2-C with respect to the single agent: (C) 4-agent, CT-graph; (D) 12-agent, CT-graph; (E) 4-agent, Minigrid; (F) 12-agent, Minigrid. (G, H, I, J) The time advantage (TA), i.e., the speedup of L2D2-C with respect to the single agent, expressed as a ratio of time-to-performance: (G) 4-agent, CT-graph; (H) 12-agent, CT-graph; (I) 4-agent, Minigrid; (J) 12-agent, Minigrid.}
    \label{fig:L2D2-C_comparison}
\end{figure}
Our initial tests aim to assess the impact on performance with an increase in the number of agents in a LL setting. In the CT-graph, 16 tasks are learned sequentially over 64 learning slots per agent. One learning slot corresponds to continuous training over one task for 12800 steps. In the Minigrid, 3 tasks are learned over 12 learning slots of 102400 steps each. The instant cumulative return (ICR) (Section \ref{sec:knowledge}) is used as the performance metric that represents the sum of returns across all tasks, i.e., $p(\mu)/ne$ (Eq.\ \ref{eq:p}). The results in Figure \ref{fig:L2D2-C_comparison} are the mean of 5 seed runs with the shades denoting the 95\% confidence interval, computed following the procedure in \cite{colas2018many}.

As the learning takes place, one evaluation agent (EA) was deployed to run a continuous assessment of L2D2-C by continuously cycling through the 16 tasks (CT-graph) and 3 tasks (Minigrid). As each agent runs independently from the others, including the EA, there is no direct mapping between evaluation checkpoints and training steps, as each agent could run at a slightly different speed. However, as the EA runs at a constant speed, the evaluation blocks can be interpreted as elapsed time. The graph indicates that the main advantage of L2D2-C is visible with a higher number of tasks (16 in the CT-graph), while such an advantage is reduced when learning the 3-task Minigrid. Nevertheless, the 12-agent L2D2-C appears immune from the local minima that affect the single agent and the 4-agent L2D2-C (Fig.\ \ref{fig:L2D2-C_comparison}(B). The second and third row of Fig.\ \ref{fig:L2D2-C_comparison} shows how much more performance is scored by L2D2-C with respect to the single agent (Panels C to F) and how much faster is L2D2-C to reach a determined level of performance (Panels G to J). From the second row of Fig.\ \ref{fig:L2D2-C_comparison}, the 4 and 12-agent L2D2-C have a significant performance advantage during the early stages of training: for the 16-task experiment, such an advantage peaks at levels that are comparable to the number of agents, i.e., approximately $n$ times performance (where $n$ is the number of agents), but decreases progressively with time as the single agent, slowly, learns all available tasks. The third row of Fig.\ \ref{fig:L2D2-C_comparison} shows that the speed advantage is also significant in the 16-task experiment and mostly exceeding a factor $n$ for target performances above $\sim 30\%$, but not so for the 3-task experiment in which the low number of tasks per agent appears to have an impact on efficiency. 

In addition to the experiments in Fig.\ \ref{fig:L2D2-C_comparison}, we run tests with the following agents/tasks numbers: 2/2, 4/4, 8/8, 4/16, 16/16, and 32/32. As these runs were executed with one seed, we report the results in the Appendix \ref{sec:additional} in Table \ref{tab:timeto} and Fig.\ \ref{fig:32x32} for the 32-tasks/32-agents run as supporting experiments. The additional runs confirm the learning dynamics of Fig.\ \ref{fig:L2D2-C_comparison} and indicate similar learning trends with up to 32 agents and 32 tasks.

\subsection{L2D2-C versus DD-PPO, IMPALA, PPO, PPO+EWC}\vspace{-3pt}
\label{sec:baselines}

We assessed the advantage of introducing LL dynamics to distributed RL, and in particular the performance of L2D2-C with respect to established distributed RL baselines, DD-PPO, IMPALA, plus two additional references PPO and PPO+EWC. Rather than measuring the ICR metric that is implemented only for L2D2-C, we monitor the performance on each task individually for both L2D2-C and the baselines.
\begin{figure}[t]
    \centering
    \begin{tabular}{ccccc}    \includegraphics[width=0.95\textwidth]{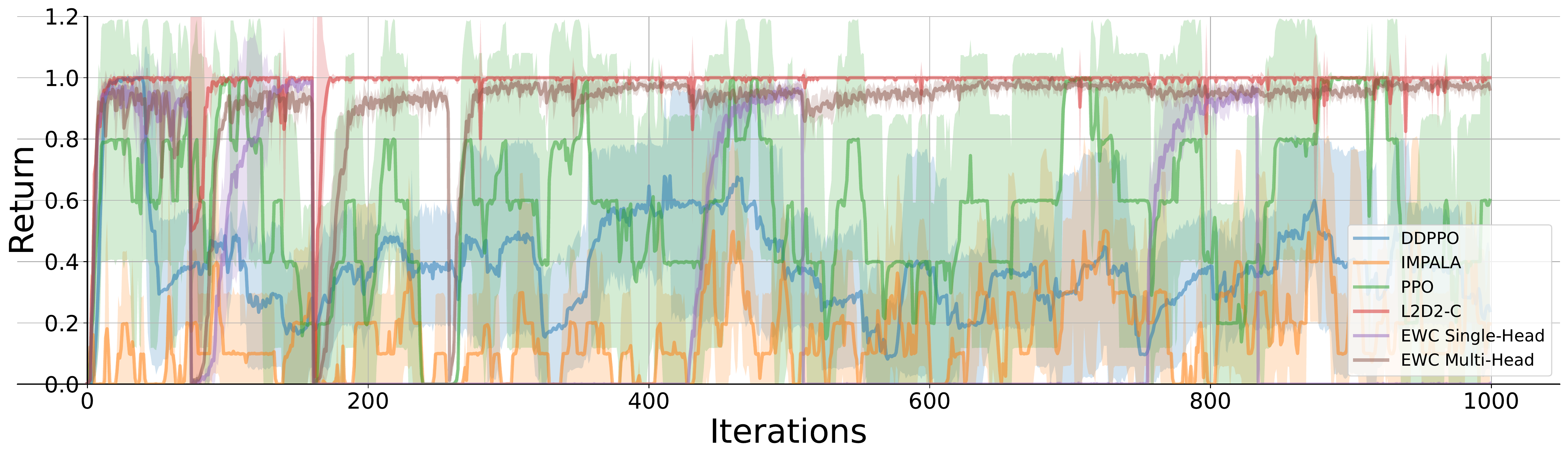}
    \end{tabular}
\caption{Training performance on single tasks through a LL curriculum: two L2D2-C agents vs.\ two DD-PPO workers, two IMPALA workers, one PPO worker, one PPO+EWC single-head and one PPO+EWC multi-head worker. The performance of one single agent is plotted for each algorithm, averaged over 5 seeds and with the shade showing the 95\% confidence interval.}
    \label{fig:baseline_comparisons}
\end{figure}
Fig.\ \ref{fig:baseline_comparisons} shows L2D2-C and the baselines sequentially learning different tasks. The task changes are visible when there is a drop in performance. It can be noted that an L2D2-C agent learns three tasks initially and then shows minor drops in performance as (1) it acquires task knowledge from the other agent and (2) it shows no forgetting thanks to the LRL masking approach. The non-LL baselines, instead, while showing steep learning curves, are unable to maintain the knowledge when switching tasks on such a sequential LL curriculum. The LL baseline PPO+EWC multi-head shows the best performance among other approaches, but such an algorithm does not have a sharing mechanism for distributed learning.

\subsection{Robustness in front of connection drops}\vspace{-3pt}
\label{sec:robustness}

\begin{figure}[t]
    \centering
    \begin{tabular}{cc}
        \includegraphics[width=0.40\textwidth]{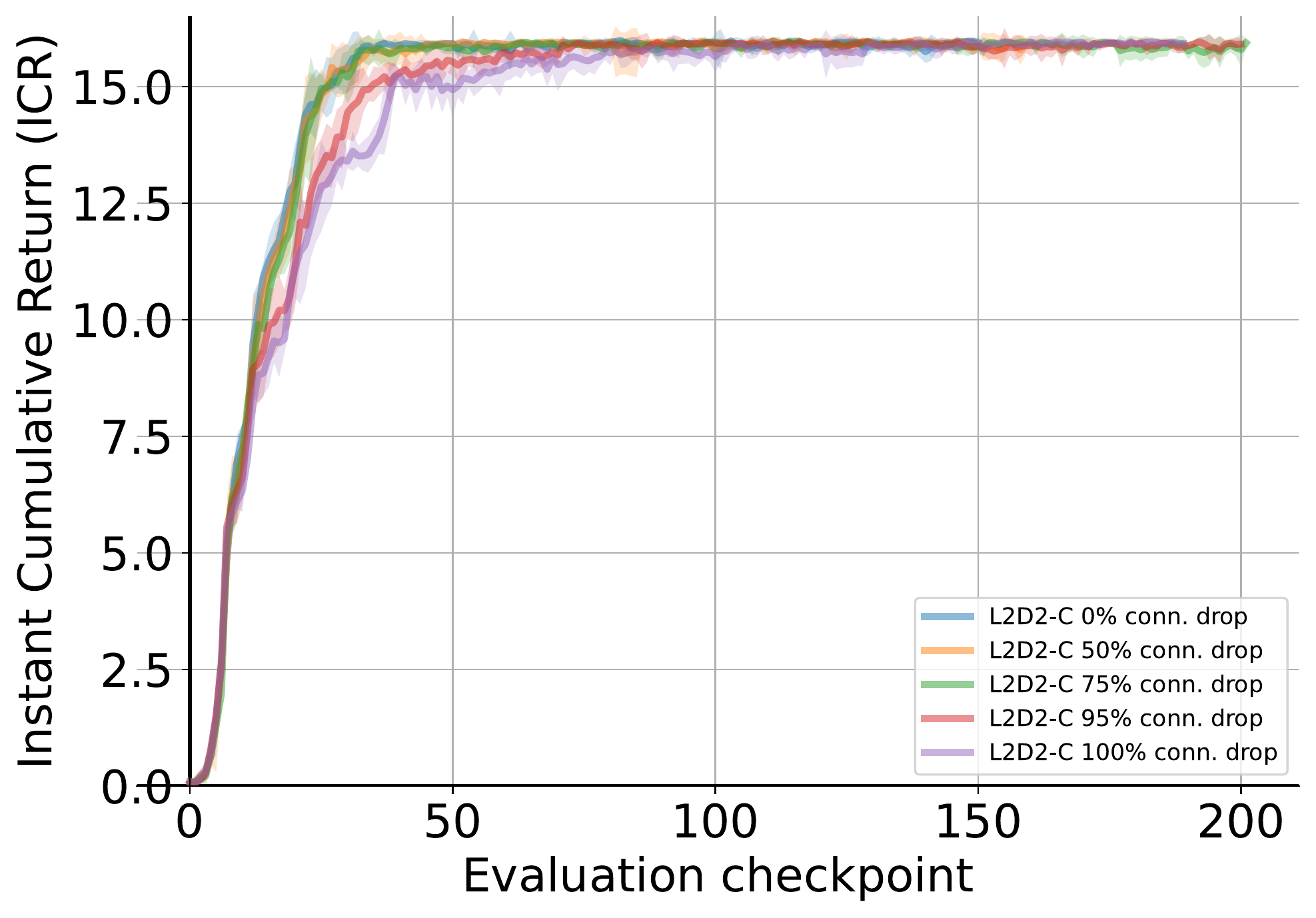} &
        \includegraphics[width=0.40\textwidth]{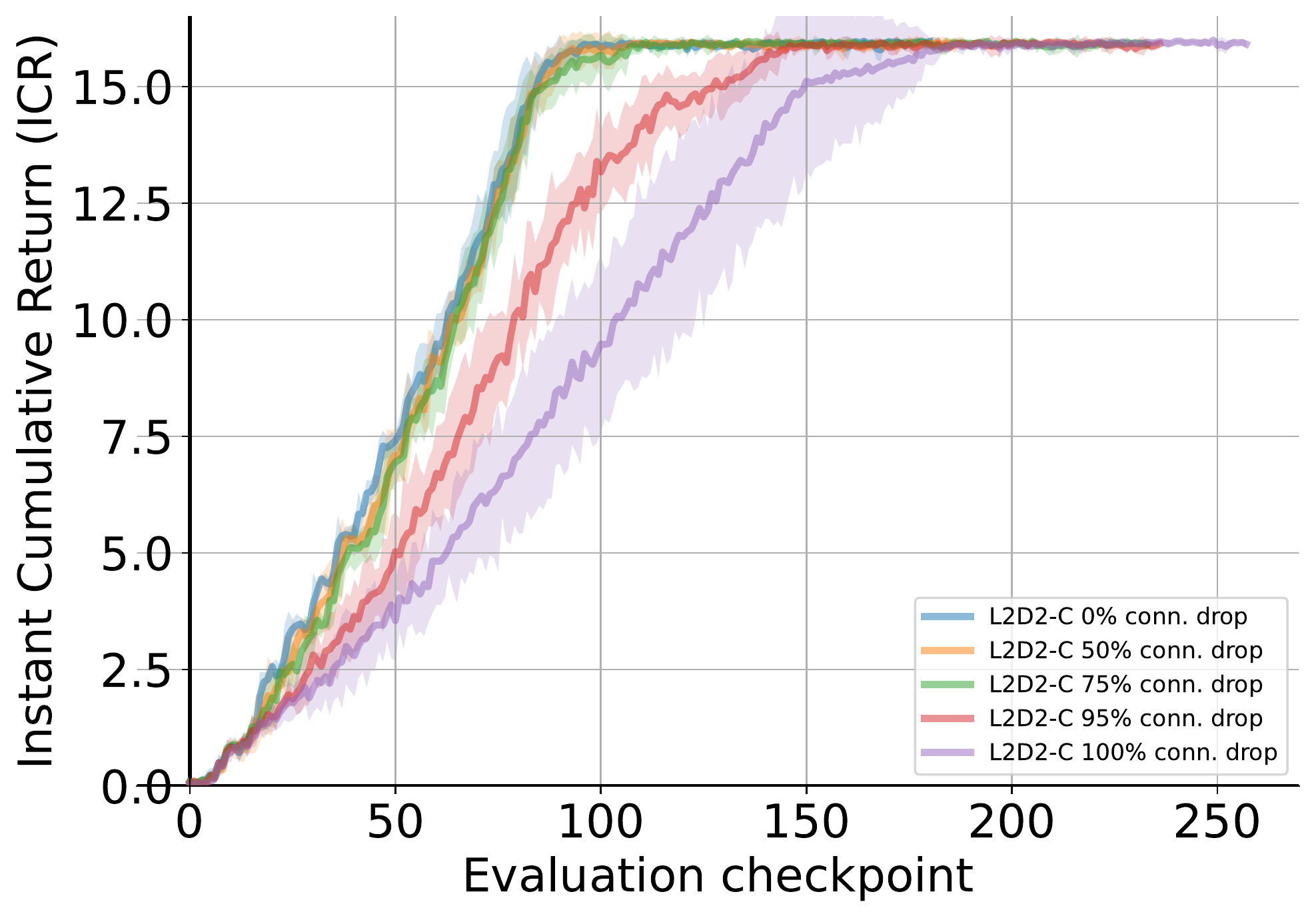}\\
        (A)&(B)\\
    \end{tabular}
    \caption{Instant cumulative reward (ICR) of L2D2-C with different connection drop probabilities for a 16-task, 12-agent CT-graph experiment (average and 95\% confidence interval over 5 seeds). (A) One evaluation agent monitors all agents. (B) One evaluation agent monitors one random agent only.}
    \label{fig:commdrop}
\end{figure}

We introduced a stochastic mechanism to simulate connection drops: with varying probability, Step 1 in the communication protocol (issuing of IDQ as outlined in Section \ref{sec:whatwhen}) is canceled. Since IDQ issuing is the first step in the communication chain, canceling this first step has the same effect as dropping communication at any following step. We use the evaluation agent in two different settings: to monitor all agents, and to monitor one single random agent. The evaluation agent is not affected by connection drops, it cannot learn, it cannot share and its presence does not affect the performance of L2D2-C. %chose, however, to maintain 100\% communication with the evaluation agent, which is not affecting the performance of the system but is capable of joining knowledge from all the agents in the collective. 
Fig.\ \ref{fig:commdrop} shows that even high levels of connection drops fail to affect the performance of the system significantly until communication is completely stopped.

\begin{figure}[t]
    \centering
    \begin{tabular}{ccc}
        \includegraphics[width=0.3\textwidth]{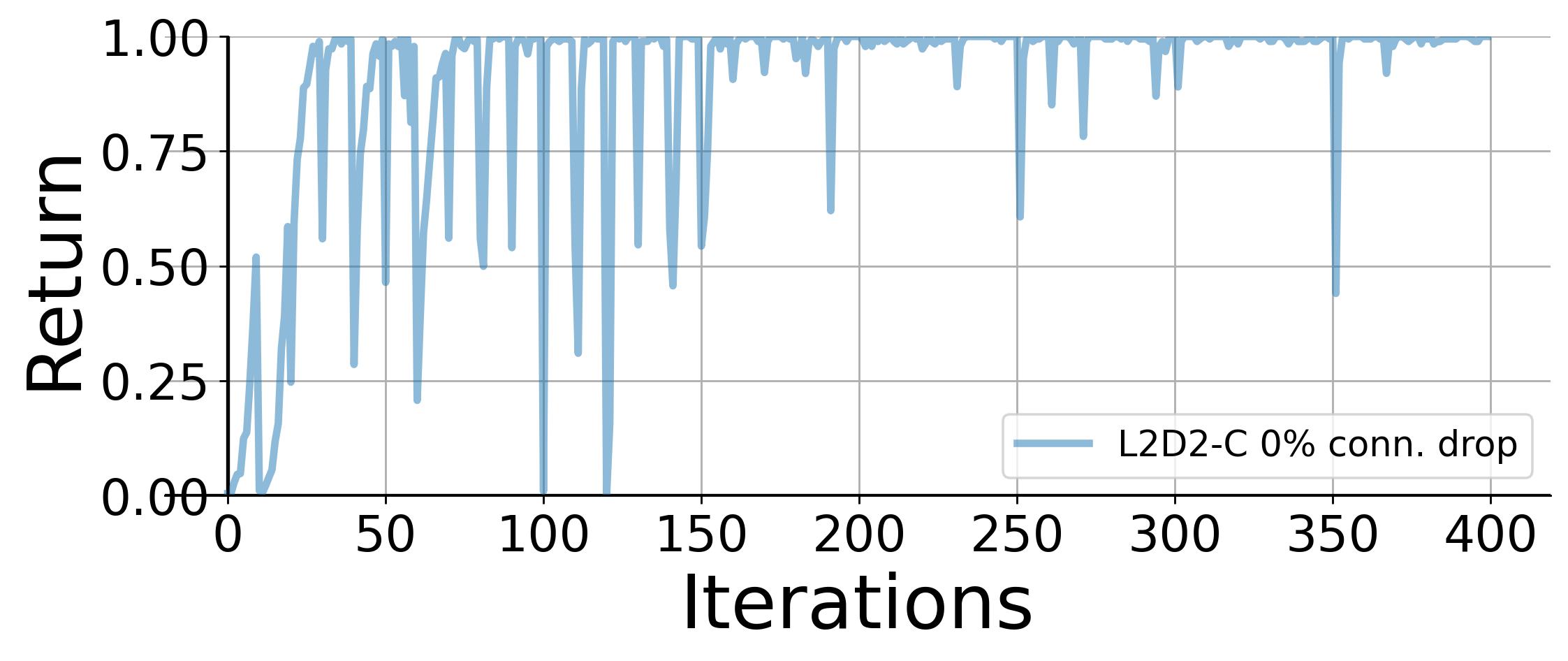} &
        \includegraphics[width=0.3\textwidth]{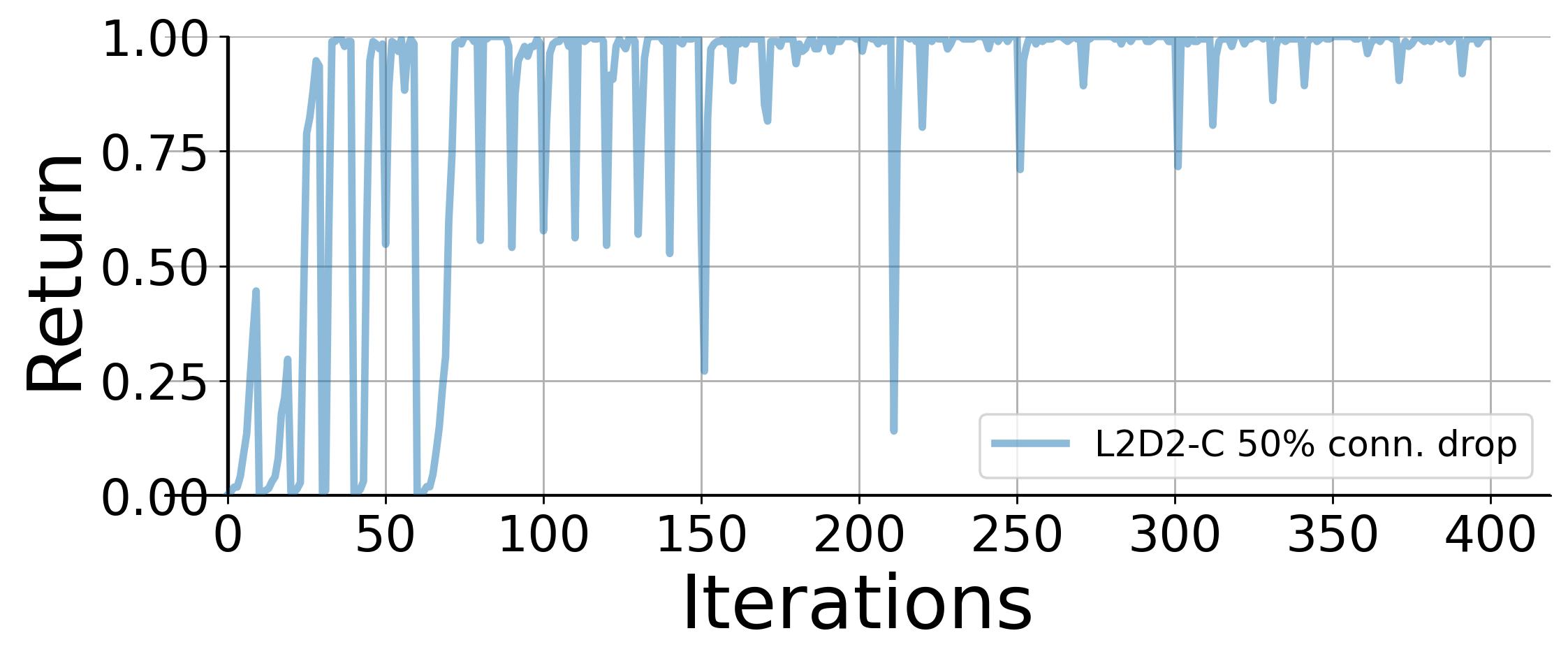} &
        \includegraphics[width=0.3\textwidth]{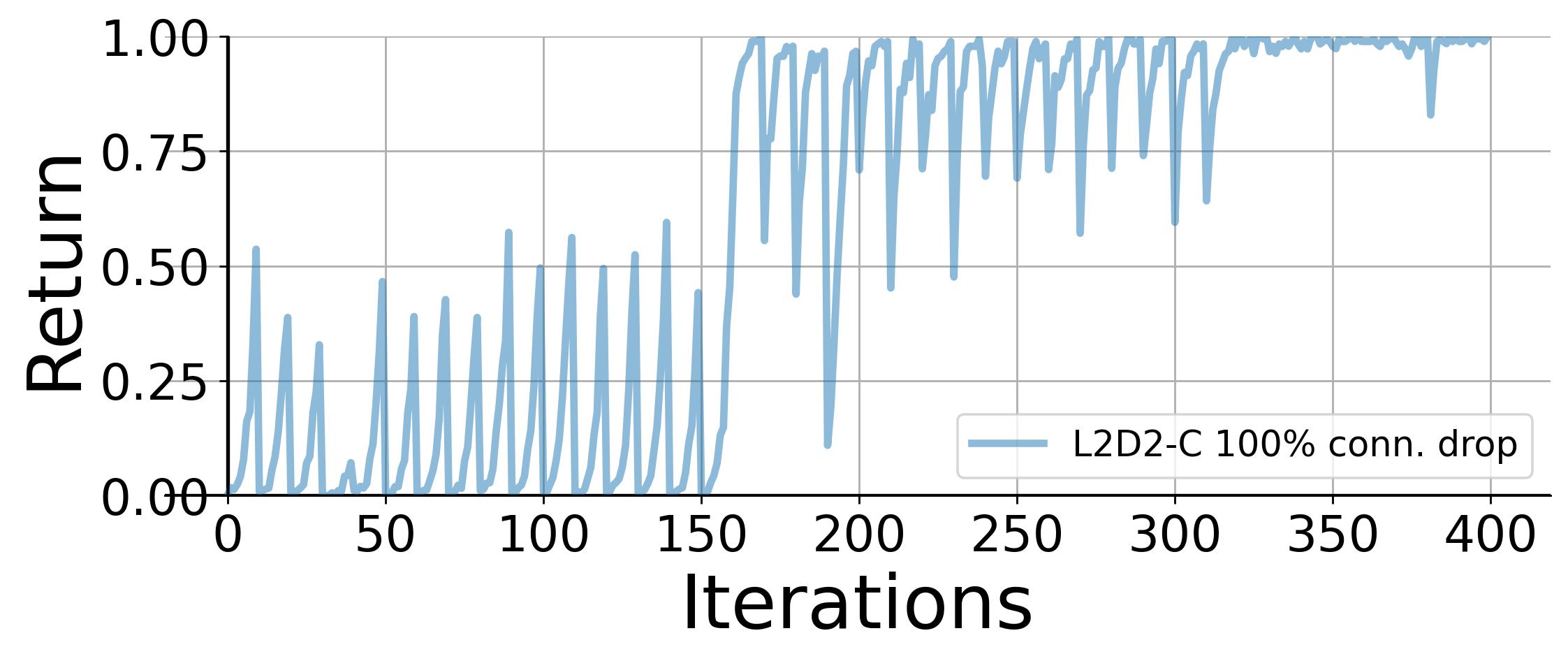}\\
        (A)&(B)&(C)\\
    \end{tabular}
    \caption{Examples of training plots for one agent with different levels of connection drops. Each task is learned for 10 iterations, equivalent to 12,800 steps. To improve readability, we modified the curriculum for this specific agent to undertake tasks sequentially from 1 to 16, and repeat that sequence.  (A) Full communication (no drops). The upwards trend from about half performance shows that the agent receives masks and then improves on them. (B) 50\% connection drops. (C) No communication. The masks now are trained partially when tasks are seen for the first time (iterations 1 to 160). The partially trained masks are used to continue training after iteration 160. No communication means that this agent is learning ``on its own''.}
    \label{fig:trainingPlots}
\end{figure}

Where does such robustness come from? We identified two reasons. Firstly, when a connection is dropped, the agent continues to learn and attempts to communicate again sometime later (when the conditions for communication occur again). In other words, the agent gives up temporarily, uses the time to learn, and tries again later. Thus, we measured that the number of transferred masks (Table \ref{tab:sharedMessages}) does not decrease linearly with the probability of connection drops, and goes to zero only when communication is completely stopped. Detailed scatter plots of data exchange are reported in the appendix (Sec.\ \ref{sec:additional}, Fig.\ \ref{fig:scatter}).  A second reason for robustness is that, even when communication is very unlikely or not possible at all, each agent is unhindered in their lifelong learning dynamics. Thus, many agents learning different tasks will collectively acquire more knowledge, which can then be shared the moment communication is restored. To provide a visual insight into learning for a single agent, Fig.\ \ref{fig:trainingPlots} shows how a single agent learns a sequence of tasks when has full communication, 50\%, and 100\% connection drops.

\begin{table}[t]
\small
    \centering
    \begin{tabular}{|l|r|r|r|r|r|}\hline
    Type of message & \multicolumn{5}{|c|}{Average nr.\ sent (5 seeds)}\\\hline
    & 0\% drop & 50\% drop & 75\% drop & 95\% drop & 100\% drop\\\hline
     ID queries (IDQ)    & 2705.98 & 1356.217 & 674.47 & 136.45 & 0 \\\hline
     Query responses (QR)& 187.23  & 95.25  & 51.33 & 18.82  & 0 \\\hline
     Mask requests (MR)  & 28.40    & 27.45   & 25.03  & 15.88  & 0 \\\hline
     Mask transfers (MTR)& 28.40    & 27.45   & 25.03  & 15.88  & 0 \\\hline
    \end{tabular}
    \caption{Average number of messages sent per type per agent with different probabilities of connection drops. It can be noted that even with high probabilities of connection drops, the agents ``insist'' on their attempts so that the number of masks transferred decreases to a lesser degree. The data shown above corresponds to the duration of learning in the collectives depicted between evaluation checkpoints 0 to 154 Fig. \ref{fig:commdrop} (A) and 0 to 192 Fig. \ref{fig:commdrop} (B).}
    \label{tab:sharedMessages}
\end{table}

%%%%%%%%%%%%%%%%%%%%
\section{Discussion}\vspace{-5pt}
\label{sec:discussion}

In our experiments, we assumed that each agent sees a random task from a list of tasks and will learn that for a fixed duration. Under such conditions, some agents might be learning the same task at the same time, while other agents learn different tasks. The ratio of tasks/agents determines the similarity of DLRL with other approaches. For one-task/many-agents, LL is not necessary and DRL performs optimally. For many-tasks/one-agent, only LL is required. For many-tasks/many-agents a combination of DRL and LL is required, as in the proposed approach. In the experiments of Fig.\ \ref{fig:L2D2-C_comparison}, we investigated two cases. In the first case (CT-graph), we used more tasks than agents (16 tasks with 12, 4, and 1 agents) and saw a linear improvement of the L2D2-C metrics with the number of agents (Fig.\ \ref{fig:L2D2-C_comparison} (CDGH)). In the second case (Minigrid), we used fewer tasks than agents (3 tasks with 12, 4, and 1 agents) and noticed the ability of parallel search to escape local minima as the number of agents increases (Fig.\ \ref{fig:L2D2-C_comparison} (B)).

L2D2-C requires no central coordination with the exception of the assignment of the task ID, for which we assume the existence of an oracle to provide each agent with the ID for each task. Such an assumption could be unrealistic in real-world scenarios where tasks are not defined by an oracle \citep{rios2020lifelong}. Extensions of the L2D2-C framework could be considered by adding task detection capabilities to enable agents to compare tasks according to their similarities \citep{liu2022wasserstein}. Such added capabilities would affect the queries that agents exchange (steps 1 and 2 in Section \ref{sec:whatwhen}) that would therefore contain task descriptors as opposed to task IDs. 

Further analysis of the system is required to assess the overall communication load and the impact of different network topologies. The query messages (IDQ and QR, Section \ref{sec:whatwhen}) are designed to discover which agents have knowledge of a specific task, and therefore, their number grows with the square of the number of agents in a fully connected network. For larger networks, query messages could be optimized by adopting multiple stellar topologies, i.e., defining some agents as hubs. However, once the mask provider has been identified, our one-to-one mask transfer protocol ensures scalability. The impact of network topologies and the communication load was not presented in this paper as the experiments in this first study were run on a single server. However, the agent-centered L2D2-C architecture was observed to be functional across distant locations with a proof-of-concept experiment in which four agents formed an L2D2-C system across locations in two different countries.  Under such conditions, analysis of the effect of bandwidth usage and delay is essential to further assess the system. 

The current approach relies on a common backbone network shared initially by all agents, and therefore it cannot be used if agents have different networks. Other approaches, e.g., based on knowledge distillation \citep{hinton2015distilling,gou2021knowledge}, could be devised for heterogeneous agents.

Important considerations for specific applications of the system could include computational and memory costs. As in all deep network approaches, these directly relate to the size of the network. In addition, by transferring highly sparse binary masks, it is possible to significantly reduce the ratio of the size of the mask with respect to the backbone. For example, for a backbone encoded with 32-bit precision parameters, a full binary mask is  1/32 the size of the backbone. Specific applications might result in different trade-offs of performance versus computation and memory. We did not implement optimization steps in the current study.

%%%%%%%%%%%%%%%%%%%%
\section{Conclusion}\vspace{-3pt}

This paper introduced the concept of a distributed and decentralized RL system that can learn multiple tasks sequentially without forgetting thanks to lifelong learning dynamics. The system is based on the idea of agents exchanging task-specific modulating masks that are applied to a backbone network that is common to all the agents in the system. While the advantage of modulating masks is known in the literature for both supervised and reinforcement learning approaches, here we show that the isolation of task knowledge to masks can be exploited to implement a fully distributed and decentralized system that is defined simply by the interconnection of a number of identical agents. The L2D2-C system was shown to maintain LL dynamics across multiple agents and to prevent catastrophic forgetting that is typical of distributed approaches such as DD-PPO and IMPALA. The system has an increase in speed-up and performance that appears to grow linearly with the number of agents when the agent/task ratio is close to one. Finally, we observed a surprising robustness of the collective learning in front of high levels of connection drops.

\subsection*{Broader Impact}\vspace{-3pt}

The idea that reinforcement learning agents can learn sequential tasks incrementally and in collaboration with other agents contributes towards the creation of potentially more effective and ubiquitous RL systems in a variety of application scenarios such as industrial robotics, search and rescue operations, and cyber-security just to mention a few. A broad application of such systems could affect efficiency and productivity by introducing automation and independent ML-driven decision-making.  Careful considerations must be taken in all scenarios in which human supervision is reduced to ensure the system acts in alignment with regulations, safety protocols, and ethical principles.  

\subsubsection*{Acknowledgments}\vspace{-3pt}

This material is based upon work supported by the Defense Advanced Research Projects Agency (DARPA) under Contract  Contract No.\, HR00112190132 (Shared Experience Lifelong Learning).

\newpage
\bibliography{collas2023_conference}
\bibliographystyle{collas2023_conference}

\newpage
\appendix

\section{Consideration on the performance of $n$ agents}
\label{apndx:performance}

With respect to a single-agent system, in a multi-agent system, the overall computational cost increases by a factor that is proportional to the number of agents, plus a communication cost. However, the aim of deploying multiple agents is to increase the speed of learning and observe possible beneficial effects of inter-agent interaction. When comparing the performance of $n$ agents versus one, it might appear as ``unfair'', but the point of the comparison is to observe how much faster can we learn a set of tasks if $n$ agents are deployed instead of one. In fact, it is not given that $n$ agents will learn $n$ times faster than the single agent: if $n$ agents do not communicate, each agent is effectively a single agent and only the merging of their knowledge would cause an acceleration of learning.  If we define a metric “learning speed ($LS$)” as the time taken by an agent to reach a pre-determined level of performance on a given curriculum of tasks, we can distinguish the following cases: 
\begin{itemize}
    \item $LS(n) \sim LS(1)$ : $n$ agents take the same amount of time to learn the curriculum as one agent. This is the case when agents do not communicate and their knowledge is not aggregated. The cost of $n$ agents is not exploited. 
    \item $1/n\cdot LS(1) < LS(n) < LS(1)$ : $n$ agents learn faster than one agent but are less than $n$ times faster than the single agent. This is the case when agents benefit from communication, but there is a loss in efficiency. The cost of $n$ agents is somewhat compensated by improved performance.
    \item $LS(n) \sim LS(1)/n$ : $n$ agents are approximately $n$ times faster than the single agent. In this case, the increased computational cost of a factor $n$ is compensated by an equivalent improvement in performance.
    \item  $LS(n) < LS(1)/n$ : $n$ n agents are faster than $n$ times the single agent. While this may appear unlikely at first, such a situation can emerge if $n$ agents exploit parallel search to solve difficult problems, e.g., with sparse reward. In this case, the cost of $n$ agents is largely compensated by a better-than-$n$ improvement in performance.
\end{itemize}

Overall, we can conclude that the deployment of $n$ agents versus one may lead to significant gains in performance if an efficient algorithm can exploit their combined computation.

\section{Hardware and software details}
\label{sec:HWSW}

 Simulations were conducted with two main network setups: one with all agents launched on a single GPU server, with specifications described in Table \ref{tab.specs} and a second setup in which agents were launched on different servers in different locations. 

\begin{figure}
\begin{center}
\includegraphics[scale=0.4]{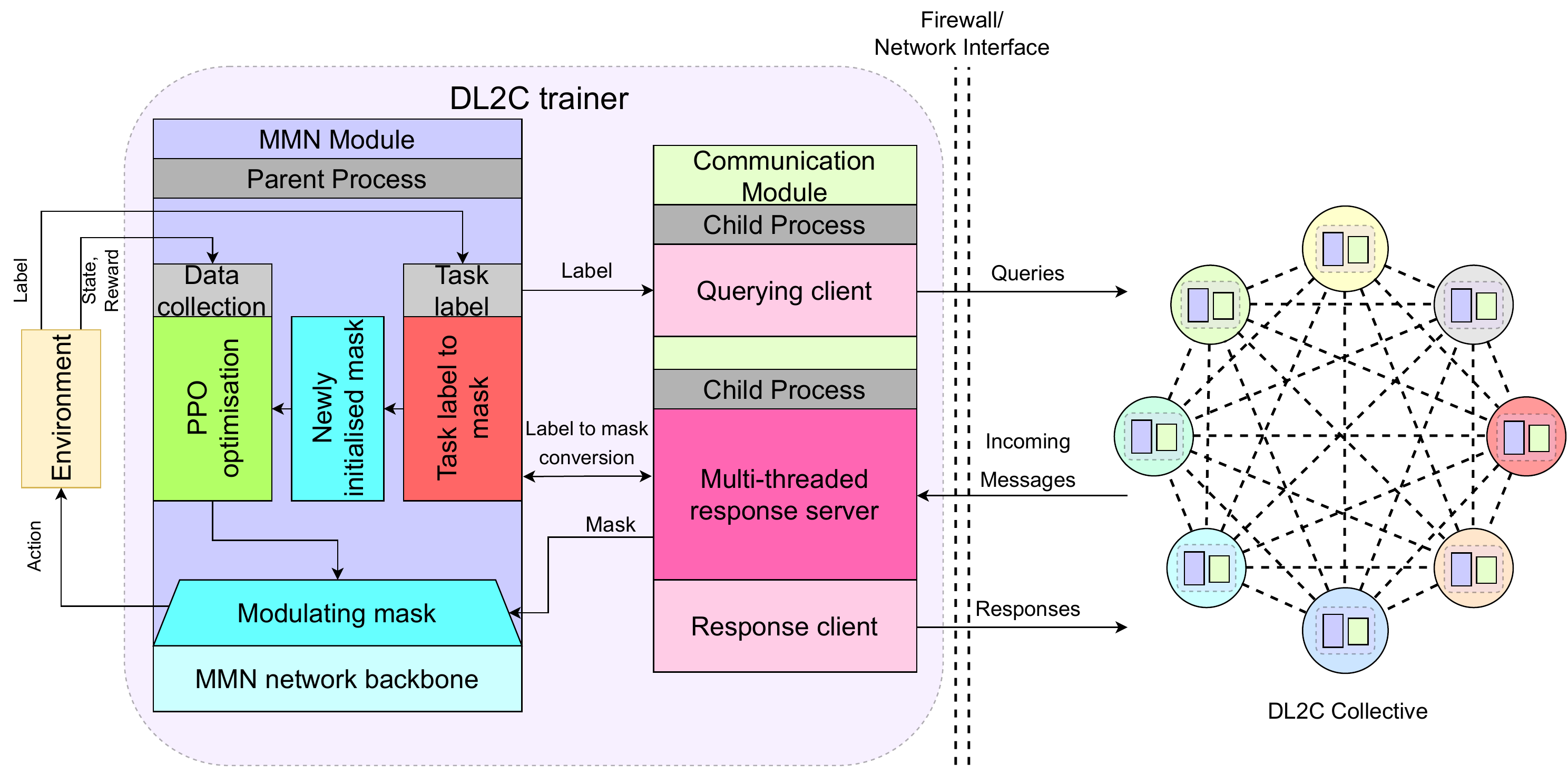}
\end{center}
\caption{System architecture of an L2D2-C agent. Collectives consist of multiple instances with a common network backbone.}
\label{fig:architecture}
\end{figure}

\begin{figure}
\begin{center}
\begin{tabular}{cc}
\includegraphics[scale=0.3]{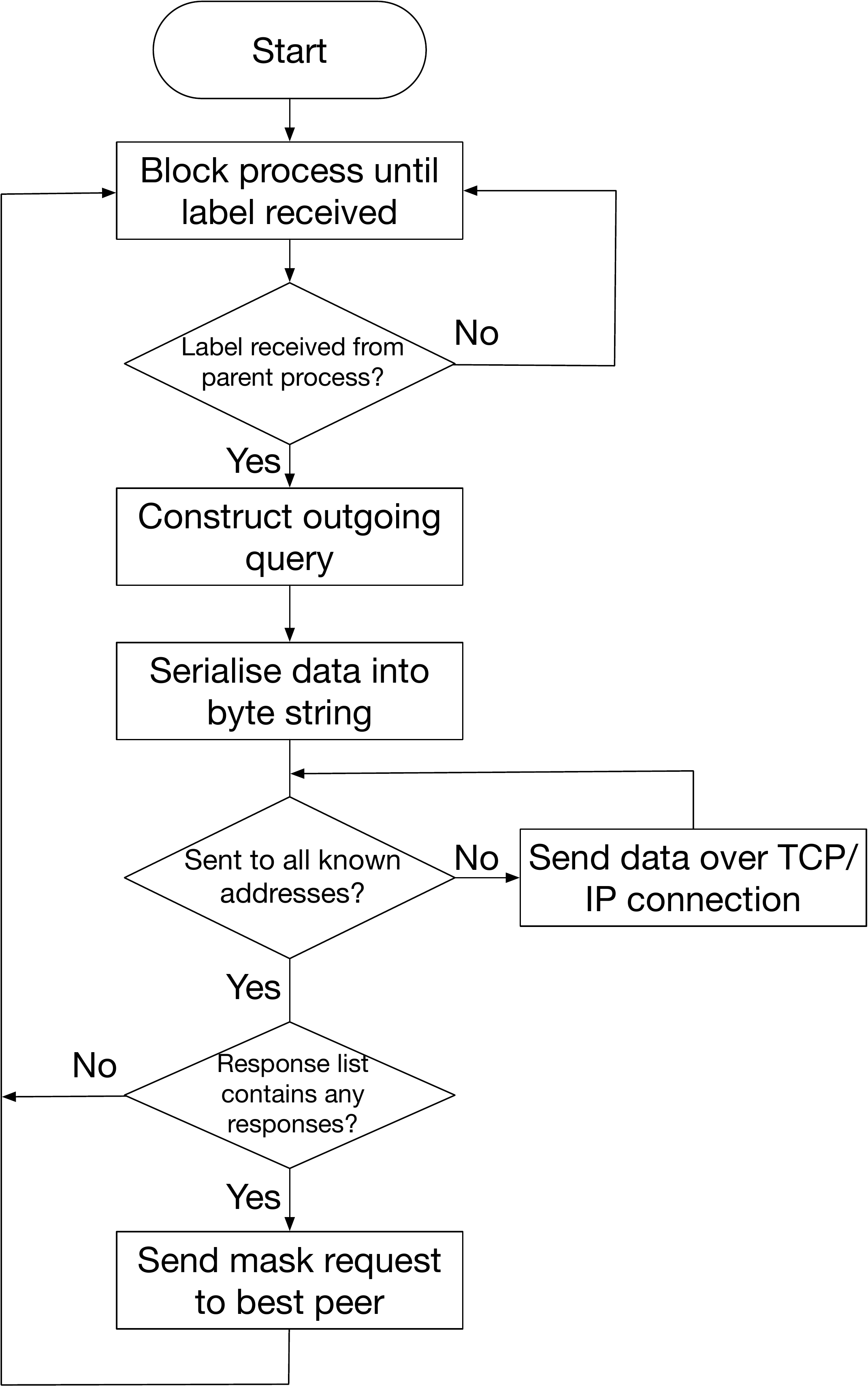} &
\includegraphics[scale=0.23]{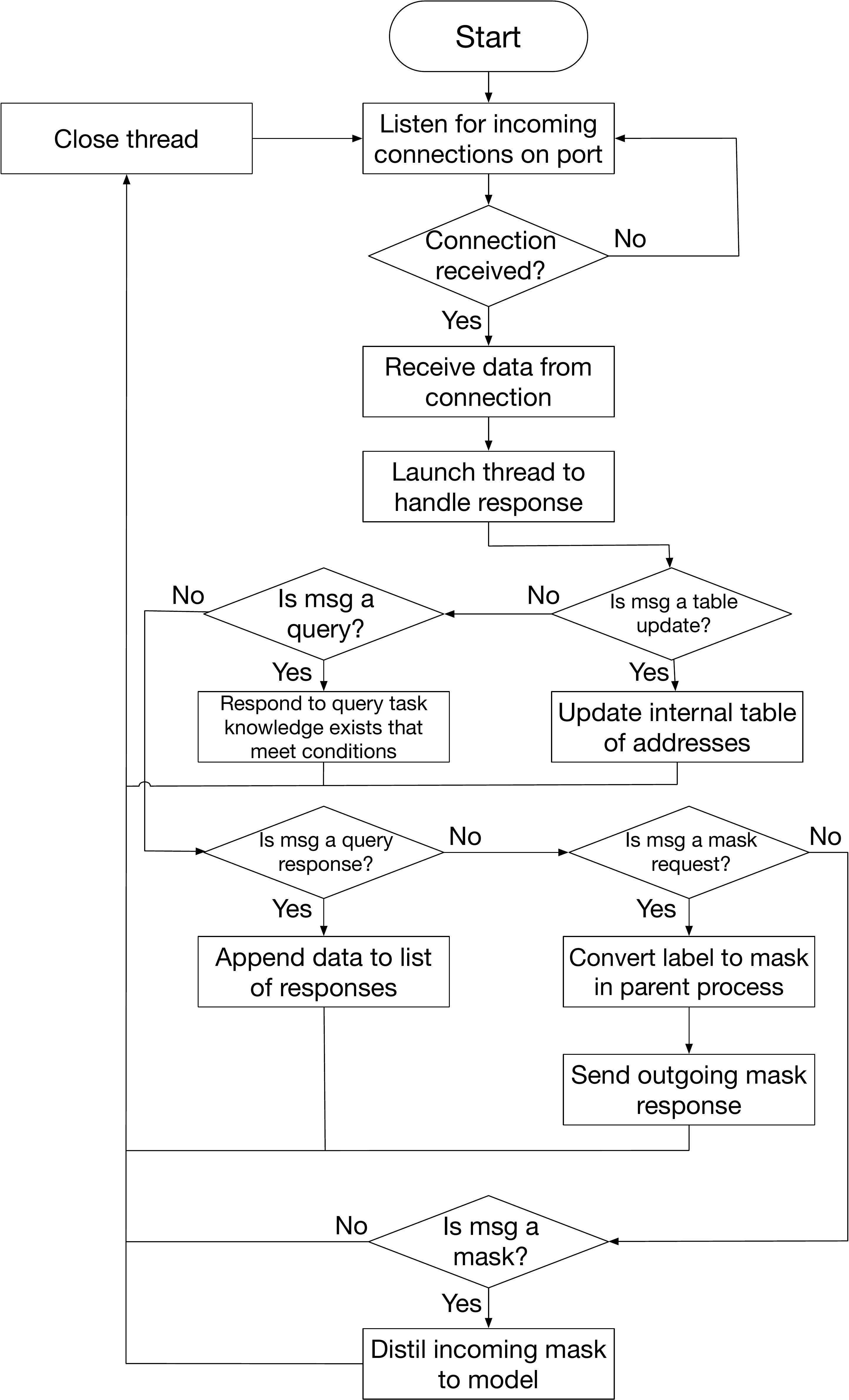}
\end{tabular}
\end{center}
\caption{Flowcharts depicting the communication protocol used in L2D2-C. Left: Client-side, Right: Server-side}
\label{fig:flowchart}
\end{figure}

Fig.\ \ref{fig:flowchart} illustrates the steps of the communication protocol (Algorithms \ref{algo1} and \ref{algo2}) in two flowcharts. 

\begin{algorithm}
    \caption{Client Protocol}
    \label{alg:commprot1}
    \begin{algorithmic}[1]
        \Require List of IP-port tuples, $C$
        \Require Current task label, $L$
        \Require Current task reward, $R$
        \Require Constant identifier for query, $I$
        \Require Own IP address, $ip_{self}$
        \Require Own port, $port_{self}$

        \While{True}
            \While{Label L not received from the main process}
                \State Wait for label L
            \EndWhile
            \State Construct query as array $\{ip_{self}, port_{self}, I, L, R\}$
            \State Serialize query into a byte string
            \For{$(ip, port)$ in $C$}
                \State Send outgoing query
            \EndFor
            \State Sleep for 0.2 seconds
            \If{Length of response list > 0}
                \State Select best agent $A_n$ using condition $A_{n} = \left\{ a\in C : 0.9 R_{n} > R_{x} \right\}$ from received responses
                \State Send mask request to $A_n$
            \EndIf
        \EndWhile
    \end{algorithmic}
    \label{algo1}
\end{algorithm}

\begin{algorithm}
    \caption{Server Protocol}
    \label{alg:commprot2}
    \begin{algorithmic}[1]
        \Require List of tuples containing IP-port pairs, C
        \Require Current task label, L
        \Require Current task reward, R
        \Require Constant identifier for query, I
        
        \State Listen for incoming connections
        \State Construct query as array $\{C_{ip}, C_{port}, I, L, R\}$
        \State Serialize query into a byte string
        \For{$(C_{ip}, C_{port})$ in $C$}
            \State Sleep for 0.2 seconds
            \If{Length of response list > 0}
                \State Select best agent $A_n$ using condition $A_{n} = \left\{ a\in C : 0.9 R_{n} > R_{x} \right\}$
                \State Send mask request to $A_n$ for $ip_n, port_n$
            \Else
                \State Pass
            \EndIf
        \EndFor
    \end{algorithmic}
    \label{algo2}
\end{algorithm}

\begin{table}
\begin{center}
\begin{tabular}{|p{3.7cm} | p{4cm}| p{4cm}|}
\hline
\multicolumn{3}{|c|}{Hardware specifications} \\
\hline
& Linux GPU Server & Windows GPU Machine\\
\hline
Architecture & x86\textunderscore{64} & x86\textunderscore{64} \\
CPU(s) & 128 & 32\\
Thread(s) per core & 2 & 2\\
Core(s) per socket & 64 & 16\\
CPU Model & AMD EPYC 7713P & AMD Threadripper Pro 3955WX\\
L1d cache & 2 MiB & 96 Kib\\
L1i cache & 2 MiB & N/A\\
L2 cache & 32 MiB & 512 Kib\\
L3 cache & 256 MiB & 64 Mib\\
CPU MHz & 1700.000 & 4000.000 \\
CPU Max MHz& 3720.7029 & 4300.000 \\
CPU Min MHz & 1500.0000 & 3900.000 \\
GPU 1 & NVIDIA A100 & NVIDIA A5000\\
GPU 2 & NVIDIA A100 & N/A\\
Interface & PCIe & PCIe\\
Memory & 256 GB ECC & 64 GB\\
GPU Memory & 40 GB ECC ($\times$2) & 24 GB ECC\\
Base GPU Clock & 765 MHz & 1170 \\
Boost GPU Clock & 1410 MHz & 1695\\
FP32 TFLOPS & 156 & 55.55 \\
FP16 TFLOPS & 312 & 111.1 \\

\hline
\end{tabular}
\vspace{8pt}
\caption{Hardware specifications for the GPU servers used to run the experiments presented in this paper. Experiments featured in Fig.\ \ref{fig:L2D2-C_comparison}, \ref{fig:commdrop} and \ref{fig:32x32} were run using the Linux-based GPU server while experiments featured in Fig.\ \ref{fig:baseline_comparisons} was run using the Windows-based machine. Experiments were run using FP16 and FP32 configurations.}
\label{tab.specs}
\end{center}
\end{table}

The code to run L2D2-C and reproduce the experiment presented in this paper is available at \codeURL.

\begin{table}
\begin{center}
\begin{tabular}{|p{4.5cm}|p{1.5cm}|}
\hline
\multicolumn{2}{|c|}{ML Packages} \\
\hline
Package & Version \\
\hline
Python & 3.9.13 \\
PyTorch & 1.13.0 \\
Gym API & 0.24.0  \\
Gym-MiniGrid & 1.1.0  \\
Gym-CTgraph & 0.1 \\
Tensorboard & 2.9.1  \\
TensorboardX & 2.5.1 \\
Pandas & 1.4.3 \\
NumPy & 1.23.1 \\
CUDA & 11.7.1 \\
cpython & 0.29.30 \\
gcc & 12.2.0 \\

\hline
\end{tabular}
\vspace{8pt}
\caption{Core packages used in a single L2D2-C agent. The core ML is based on PyTorch with CUDA. OpenAI Gym is used as the interface between the model and the reinforcement learning environment. TCP/IP communication is implemented using the built-in socket module in Python.}
\label{tab.packages}
\end{center}
\end{table}

\begin{table}
\begin{center}
\begin{tabular}{|p{5cm}|p{4cm}|}
\hline
\multicolumn{2}{|c|}{Hyper-Parameters} \\
\hline
learning rate & 0.00015\\
cl preservation & supermasks\\
number of workers & 1 \\
optimiser function & RMSprop \\
discount rate & 0.99 \\
use general advantage estimator (gae) & True \\
gae\_tau & 0.99 \\
entropy weight & 0.00015 \\
rollout length & 512 \\
optimisation epochs & 8 \\
number of mini-batches & 64 \\
ppo ratio clip & 0.1 \\
iteration log interval & 1 \\
gradient clip & 5 \\
max steps (per task) & 84480 \\
evaluation episodes & 25 \\
require task label & True \\
backbone network seed & 9157 \\
experiment seeds & 958, 959, 960, 961, 962 \\
\hline
\end{tabular}
\caption{Agent hyper-parameters. These parameters were used across all the single-server experiments shown in Fig.\ \ref{fig:baseline_comparisons}. These hyper-parameters are consistent across all experiments shown in the figure. IMPALA, DDPPO, and PPO experiments were run using the Ray API \citep{ray2023}.}\vspace{8pt}
\label{tab.baseline_hyperparameters_l2d2-c}
\end{center}
\end{table}

\begin{table}
\begin{center}
\begin{tabular}{|p{5cm}|p{2.5cm}|p{2.5cm}|p{2.5cm}|}
\hline
Hyper-Parameter & CT-graph & Minigrid & CT-graph Drop conn. \\
\hline
learning rate & 0.00015 & 0.00015 & 0.00015\\
cl preservation & supermasks & supermasks & supermasks\\
number of workers & 4 & 4 & 4\\
optimiser function & RMSprop & RMSprop & RMSprop\\
discount rate & 0.99 & 0.99 & 0.99 \\
use general advantage estimator (gae) & True & True & True \\
gae\_tau & 0.99 & 0.99 & 0.99 \\
entropy weight & 0.00015 & 0.1 & 0.00015\\
rollout length & 128 & 128 & 320 \\
optimisation epochs & 8 & 8 & 8 \\
number of mini-batches & 64 & 64 & 64 \\
ppo ratio clip & 0.1 & 0.1 & 0.1 \\
iteration log interval & 1 & 1 & 1 \\
gradient clip & 5 & 5 & 5 \\
max steps (per task) & 12800 & 102400 & 12800\\
evaluation episodes & 25 & 5 & 25 \\
require task label & True & True & True \\
backbone network seed & 9157 & 9157 & 9157 \\
\hline
\end{tabular}
\caption{Agent hyper-parameters. These parameters were used across all the single-server experiments shown in Fig.\ \ref{fig:L2D2-C_comparison} and \ref{fig:commdrop}. Fig.\ \ref{fig:commdrop} specifically uses the connection drops version of the CT-graph hyper-parameters.}\vspace{8pt}
\label{tab.hyperparameters}
\end{center}
\end{table}

\begin{table}
\begin{center}
\begin{tabular}{|p{1cm}|p{2cm}|p{2cm}|p{2cm}|p{2cm}|p{2cm}|}
\hline
Agents & seed 1 & seed 2 & seed 3 & seed 4 & seed 5 \\
\hline

1 & 9158 & 6302 & 8946 & 5036 & 7687 \\
2 & 9159 & 1902 & 4693 & 8814 & 1029 \\
3 & 9160 & 4446 & 3519 & 2851 & 4641 \\
4 & 9161 & 9575 & 6652 & 4719 & 7122 \\
5 & 9162 & 1954 & 6613 & 3672 & 6470 \\
6 & 9163 & 3972 & 4197 & 3523 & 5614 \\
7 & 9164 & 5761 & 8640 & 6978 & 4687 \\
8 & 9165 & 8004 & 6738 & 1399 & 8024 \\
9 & 9166 & 3993 & 8248 & 5952 & 3218 \\
10 & 9167 & 6553 & 7423 & 9744 & 8224 \\
11 & 9168 & 8805 & 9725 & 3633 & 2047 \\
12 & 9169 & 5066 & 8760 & 4131 & 6219 \\
\hline
0 & 9157 & 9802 & 9822 & 2211 & 1911 \\
13 & 9157 & 9802 & 9822 & 2211 & 1911 \\
\hline
\end{tabular}
\caption{Seeds used across each agent. These remained consistent throughout each experiment. Global evaluation agents correspond to agent 0 (evaluate the entire collective), and local evaluation agents correspond to agent 13 (evaluate a singular agent from the collective). A single L2D2-C agent corresponds to agent 1, and L2D2-C collectives are comprised of subsets of these agents (i.e., L2D2-C 4 agents consist of agents 1-4 while L2D2-C 12 agents consist of agents 1-12).}\vspace{8pt}
\label{tab.seeds}
\end{center}
\end{table}

\begin{table}
\begin{center}
\begin{tabular}{|p{4.9cm}||p{3.8cm}|}
\hline
Hyper-Parameter & Depth 2 16-task CT-graph \\
\hline
general seed & 3\\
tree depth & 2\\
branching factor & 2 \\
wait probability & 0.0 \\
high reward value & 1.0 \\
fail reward value & 0  \\
stochastic sampling & false  \\
reward standard deviation & 0.1  \\
min static reward episodes & 0 \\
max static reward episodes & 0 \\
reward distribution & needle in haystack \\
MDP decision states & true  \\
MDP wait states & true  \\
wait states & $[2,8]$ \\
decision states & $[9,11]$  \\
graph ends & $[12,15]$ \\
image dataset seed(s) & 1-4  \\
1D format & false  \\
image dataset seed & 1 \\
number of images & 16  \\
noise on images on read & 0 \\
small rotation on read & 1 \\
\hline
\end{tabular}
\caption{CT-graph environment parameters. We use 5 seed variations of a depth 2 4-task CT-graph to achieve the 16-task CT-graph for our experiments. We use the depth 2 16-task CT-graph for Fig.\ \ref{fig:L2D2-C_comparison} (A, C, D, G, H) and \ref{fig:commdrop}. %We use the depth 4 16-task CT-graph for figure \ref{fig:intratask}.
}\vspace{8pt}
\label{tab.ctparams}
\end{center}
\end{table}

\begin{table}
\begin{center}
\begin{tabular}{|p{4cm}||p{6cm}|}
\hline
Hyper-Parameter & 3-task Minigrid \\
\hline
tasks & MiniGrid-SimpleCrossingS9N1-v0, MiniGrid-SimpleCrossingS9N2-v0, MiniGrid-SimpleCrossingS9N3-v0 \\
one hot & true \\
label dimensions & 3 \\
action dimensions & 3 \\
seeds & 860, 860, 860 \\
\hline
\end{tabular}
\caption{Minigrid environment parameters. Each seed corresponds to task IDs 0-2 respectively. We use the 3-task Minigrid for Fig.\ \ref{fig:L2D2-C_comparison} (B, E, F, I, J).}\vspace{8pt}
\label{tab.mgparams}
\end{center}
\end{table}

\section{Additional results}
\label{sec:additional}

\begin{figure}
    \centering
    \includegraphics[width=0.5\textwidth]{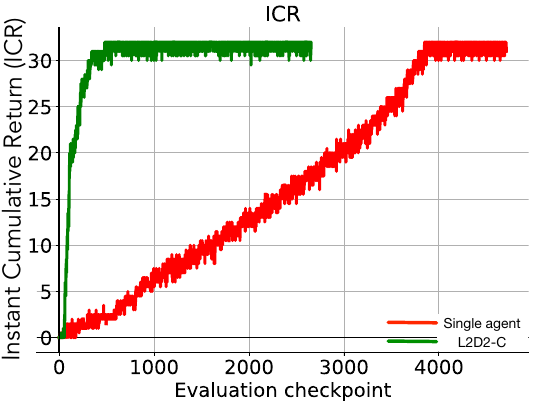}
    \caption{Instant cumulative reward (ICR) for the 32-agents/32-tasks experiment.}
    \label{fig:32x32}
\end{figure}

Table \ref{tab:timeto} reports additional metrics for single-seed experiments in which the agents/tasks ratio was used: 2/2, 4/4, 8/8, 16/16, 32/32, and 4/16. The instant cumulative reward (ICR) for the 32/32 experiment is reported in Fig.\ \ref{fig:32x32}.

\begin{figure}
    \centering
    \includegraphics[width=0.4\textwidth]{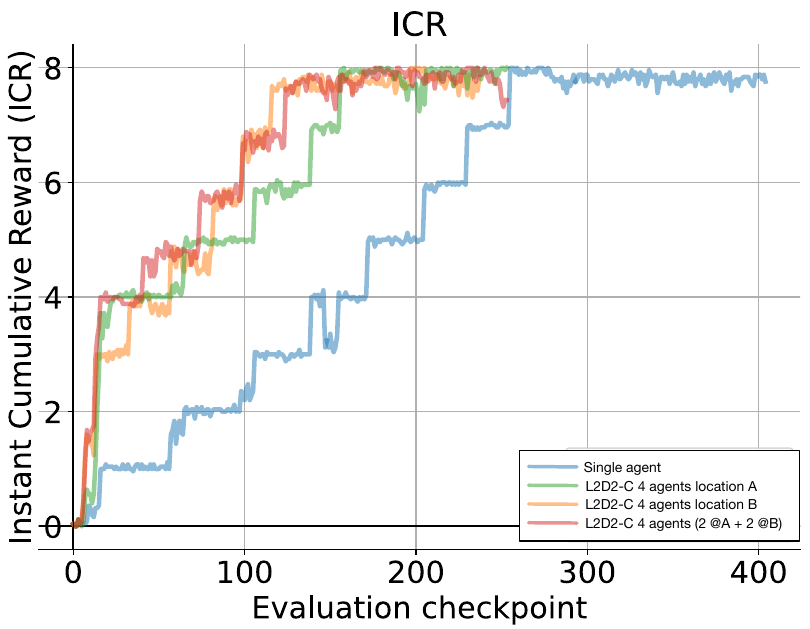}
    \caption{An L2D2-C system run across two locations in two different countries. The system across two locations does not appear to be slower than the system that ran in the same location.}
    \label{fig:2locations}
\end{figure}

Fig.\ \ref{fig:2locations} shows the performance of an L2D2-C system across two locations in different countries. This is a proof-of-concept that the system can work across locations and more statistical analysis and experiments are required. However, it indicates that four agents across two locations do not appear slower than 4 agents in the same location. 

\begin{figure}
    \centering
    \begin{tabular}{cc}
        \includegraphics[width=0.45\textwidth]{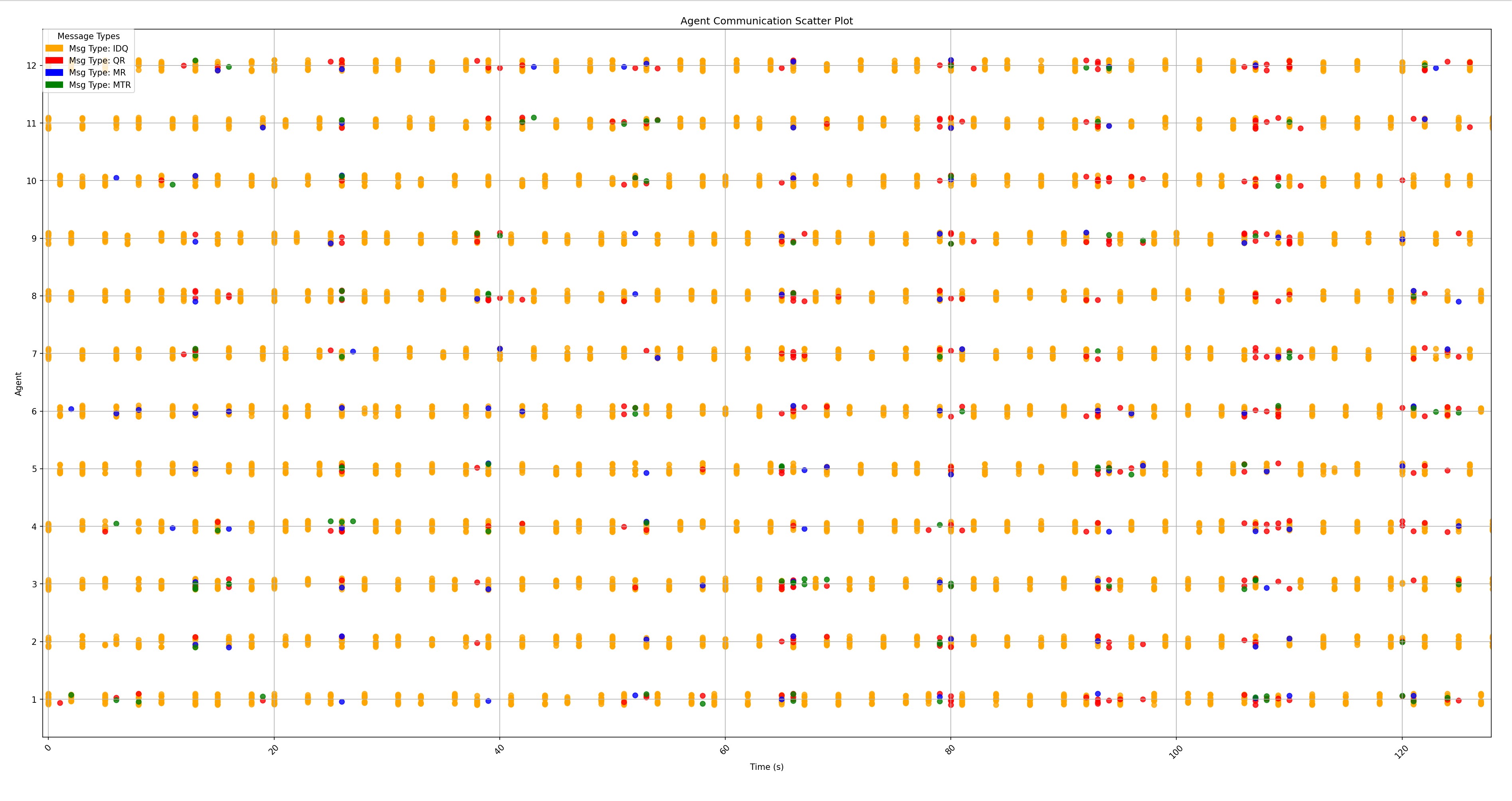} & \includegraphics[width=0.45\textwidth]{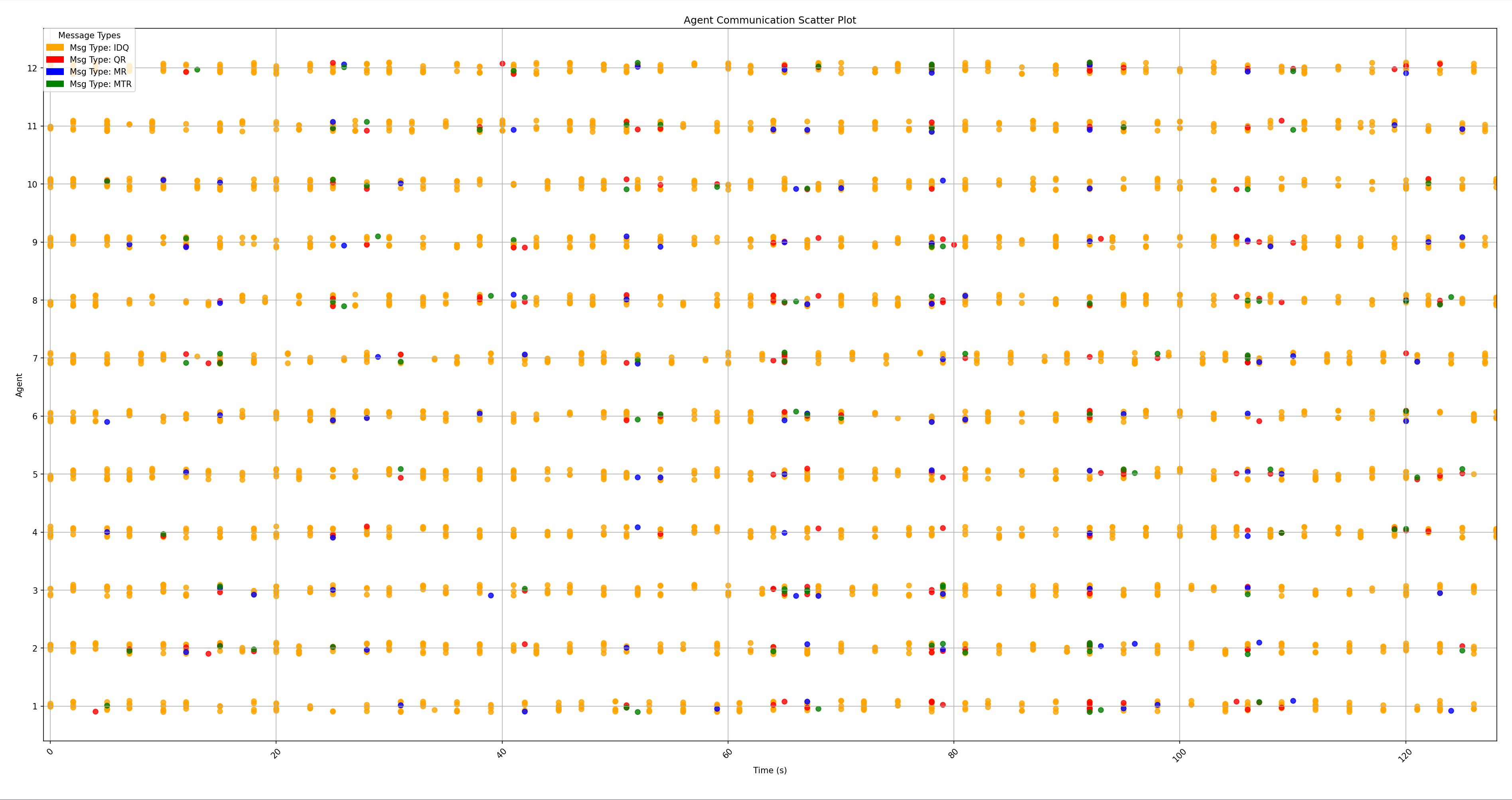} \\ 
         
        (A)&(B)\\
        \includegraphics[width=0.45\textwidth]{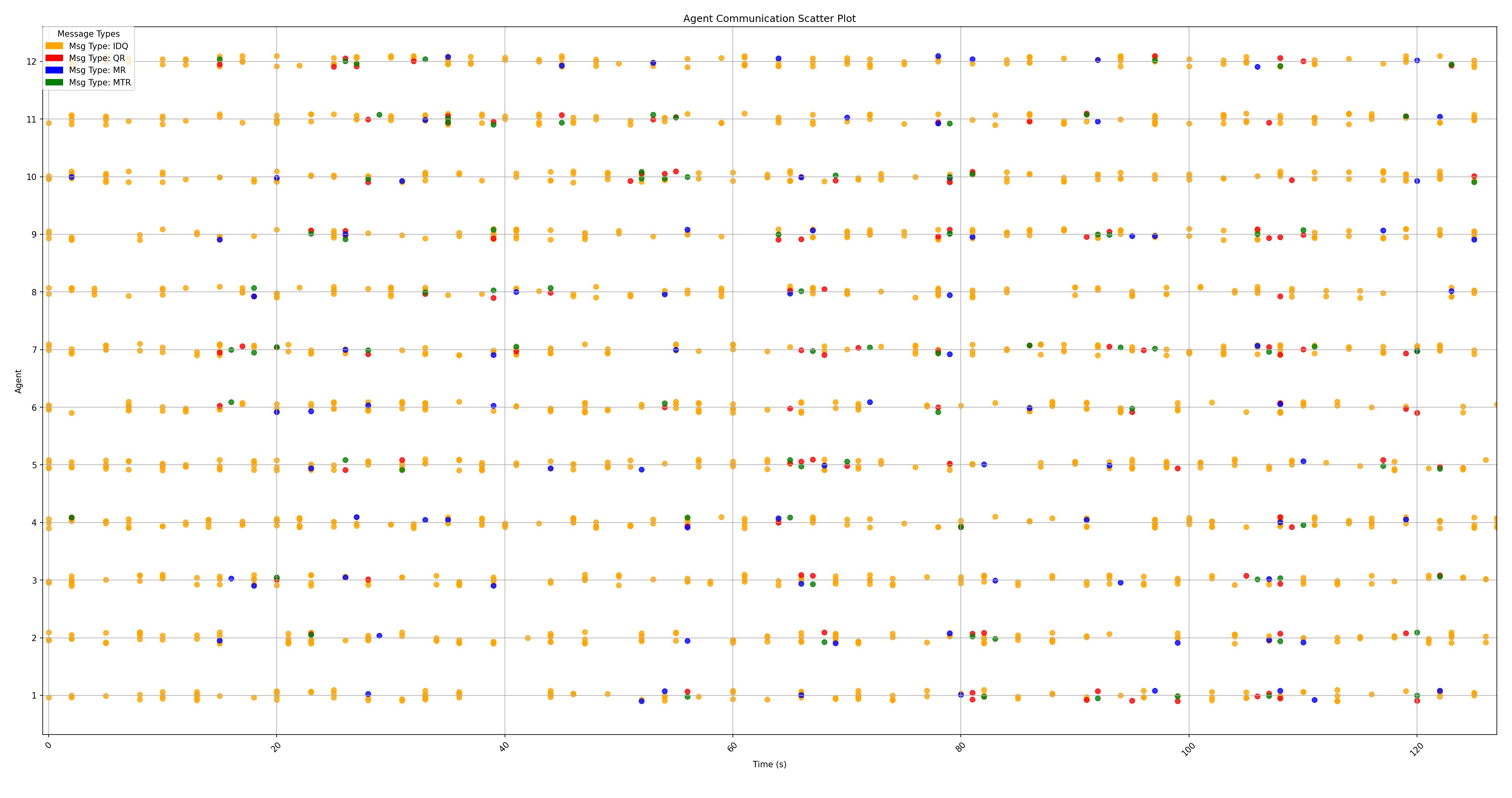} & \includegraphics[width=0.45\textwidth]{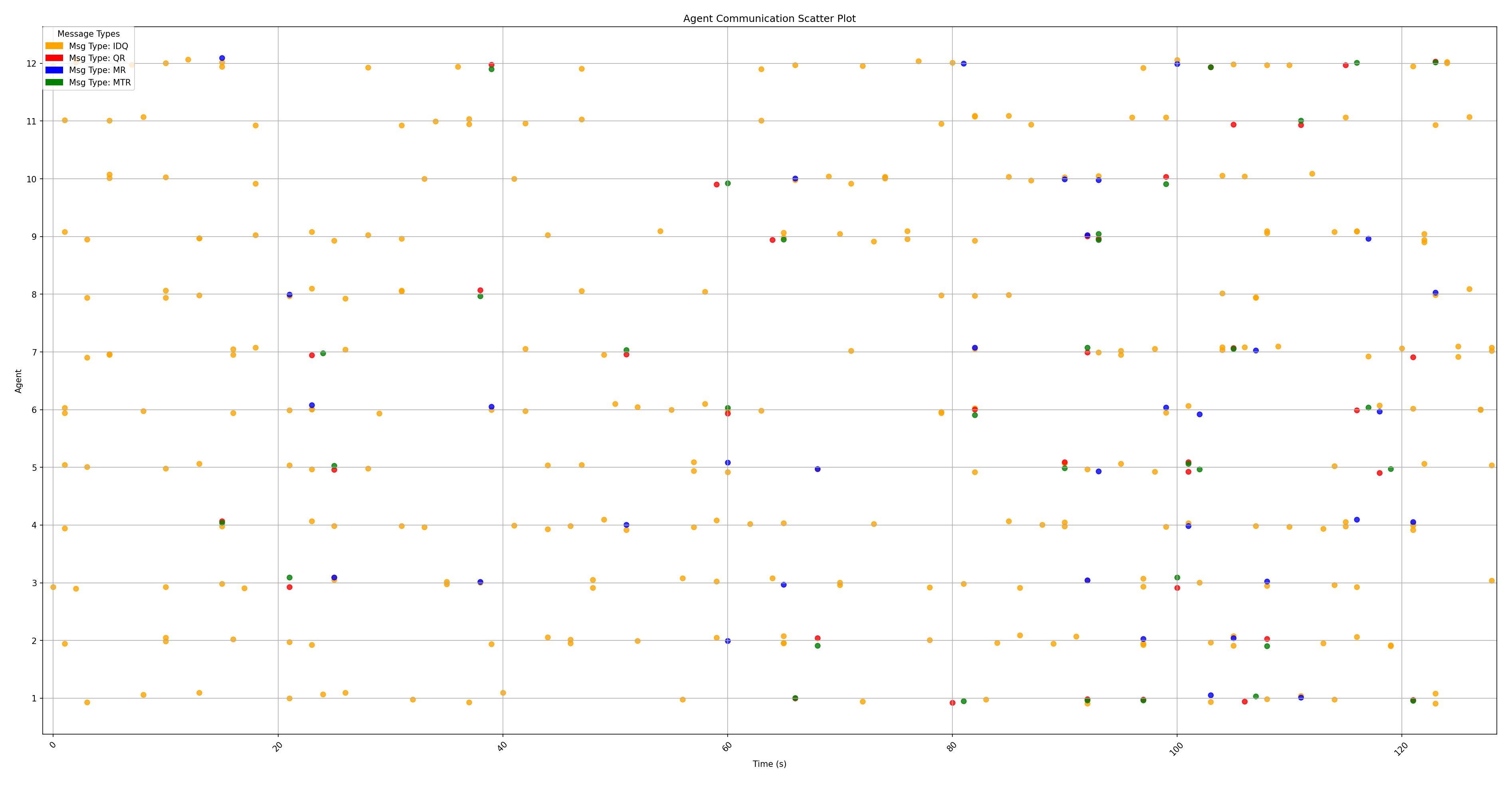}\\ 
        (C)&(D)
    \end{tabular}
        \caption{Different message types exchanged by 12 agents during runs with different connection drop probabilities. For better visibility, a time window of 120 seconds only is shown. (A) Full connection. (B) 50\% connection drop probability. (C) 75\% connection drop probability. (D) 95\% connection drop probability.}
    \label{fig:scatter}
\end{figure}

Scatter plots showing the different types of messages exchanged by all agents during 4 runs with different probabilities of connection drop are shown in Fig.\ \ref{fig:scatter}. 

\begin{table}
\centering
\small
\begin{tabular}{|c|c|c|c||c|c|c|}
\hline
\textbf{agents/tasks} & \multicolumn{3}{|c||}{\textbf{Time to X\% performance}} & \multicolumn{3}{|c|}{\textbf{Performance at X\% time}}\\
\hline
      &  \textbf{20\%} & \textbf{50\%} & \textbf{70\%} & \textbf{20\%} & \textbf{50\%} & \textbf{70\%}\\
    \hline
    2/2 (L2D2-C) & 1.7 & 27.9 & 33.1  & 6.6 & 100 & 100\\
    1/2 (singleLLAgent) & 4.1 & 33 & 57 & 0 & 50 & 48.3\\
\hline
    4/4 (L2D2-C) & 10.2 & 12.5 & 16.3 & 71.2 & 97.5 & 100.0\\
    1/4 (singleLLAgent) & 11.1 & 45.8 & 76.0 & 22.5 & 43.1 & 50.0\\
    \hline
8/8 (L2D2-C) & 5.4 & 6.9 & 16.3 & 72.7 & 98.7 & 98.5\\
    1/8 (singleLLAgent)& 22.4 & 55.8 & 79.9 & 12.9 & 32.7&57.0\\
    \hline
    \textbf{16/16 (L2D2-C)} & \textbf{2.9} & \textbf{4.5} & \textbf{8.6} & \textbf{100.0} & \textbf{100.0} & \textbf{100.0}\\
    1/16 (singleLLAgent) & 24.6& 57.5&76.9 & 12.5 & 40.6 & 56.2\\
    \hline
    \textbf{32/32 (L2D2-C)} & \textbf{1.8}& \textbf{2.7}&\textbf{4.5} & \textbf{96.8} & \textbf{100.0} & \textbf{96.8}\\
    1/32 (singleLLAgent)& 22.8 & 60.5& 79.2 & 12.5&37.5 &54.6\\
    \hline
    4/16 (L2D2-C) & 5.1 & 18.8 & 28.0 & 55.4 & 98.7 & 100\\
    1/16 (singleLLAgent)& 24.5& 55.2& 76.2& 12.5 & 37.5 & 59.6\\
    \hline
\end{tabular}
\caption{Time to X\% performance (in \%) and performance (in \%) at X\% time (with total time measured as the time required to reach 95\% performance. Due to noise affecting evaluations with few tasks, smoothing was applied: percentages are more representative for experiments with 8 or more tasks. L2D2-C 16/16 and 32/32 (in bold) were the fastest vs the single agent.}
\label{tab:timeto}
\end{table}

%%%%%%%%%%%%%%%%%%%%%%%%%%%%%%%%%%
\section{Environments}
\label{apndx:environments}
\subsection{CT-graph}
\label{apndx:env-ctgraph}
\begin{figure}
    \centering
    \begin{tabular}{cc}
    \includegraphics[width=0.45\textwidth]{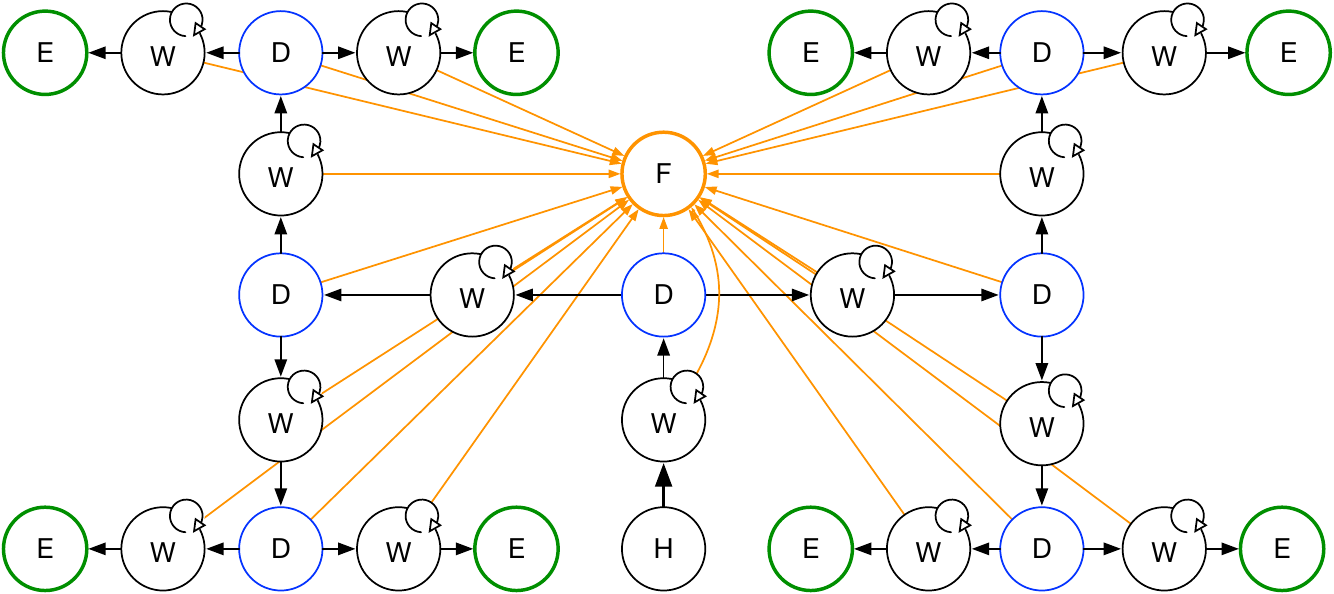}
         &     \includegraphics[width=0.21\textwidth]{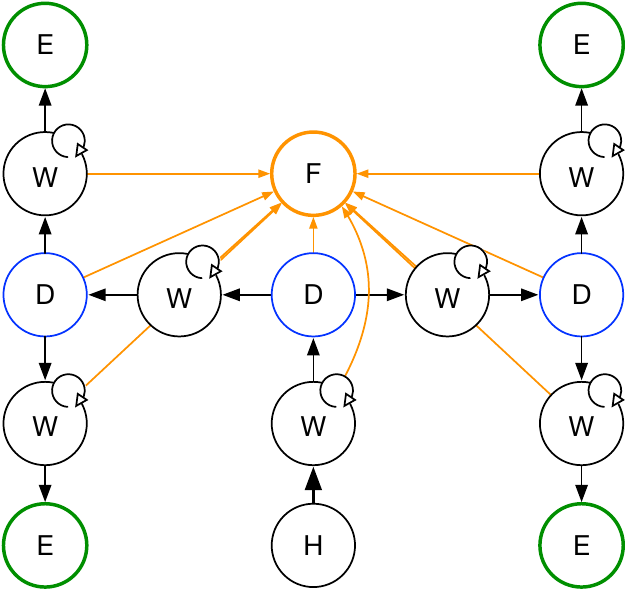}
 \\
    \end{tabular}
    \caption{Examples of CT-graph environments. States are: home (H), wait (W), decision (D), end (E), and fail (F). Three actions at W and D nodes determine the next state. (Left) A depth-3 graph with three sequential decision states (D) and reward probability $1/3^7 = 1/2187$ reward/episodes. (Right) A depth-2 graph with 4 leaf states.}
    \label{fig:ct-graphs}
\end{figure}

In the configurable tree graph (CT-graph) \citep{soltoggio2019ctgraph,soltoggio2023configurable} the environment is implemented as a graph, where each node is a state represented as a $12 \times 12$ grayscale image. As shown in Fig.\ \ref{fig:ct-graphs}, the node types are start (H), wait (W), decision (D), end/leaf (E), and fail (F). The goal of an agent is to navigate from the home state to one of the end states designated as the goal where the agent receives a reward of 1. To navigate to any end of the graph, the agent is required to perform action $a_0$ when in W states, and any other action $a_i$ with $i > 0$ in D states. The challenge is any policy that does not follow the above criteria will lead the agent to the fail state and terminate the episode without reward. 

Two parameters in the CT-graph are the branching factor $b$ that determines how many sub-trees stem from each D state, and the depth $d$ that determines how many subsequent branching node D occur before a leaf node. By setting these two parameters, different instances of the CT-graph can be created with a measurable size of the search space and reward sparsity.

\begin{figure}
    \centering
    \includegraphics[width=1\textwidth]{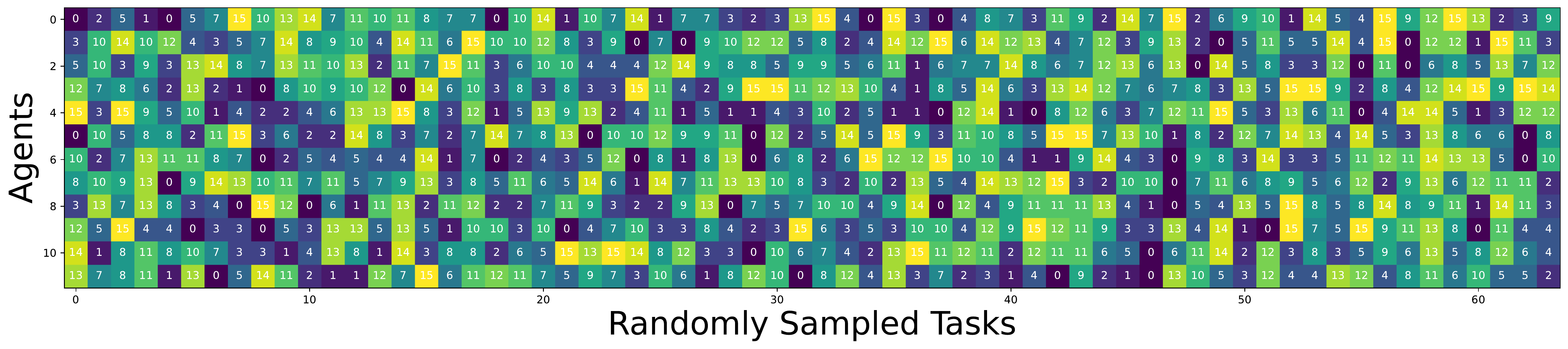} \\
    (A) \\
    \includegraphics[width=0.56\textwidth]{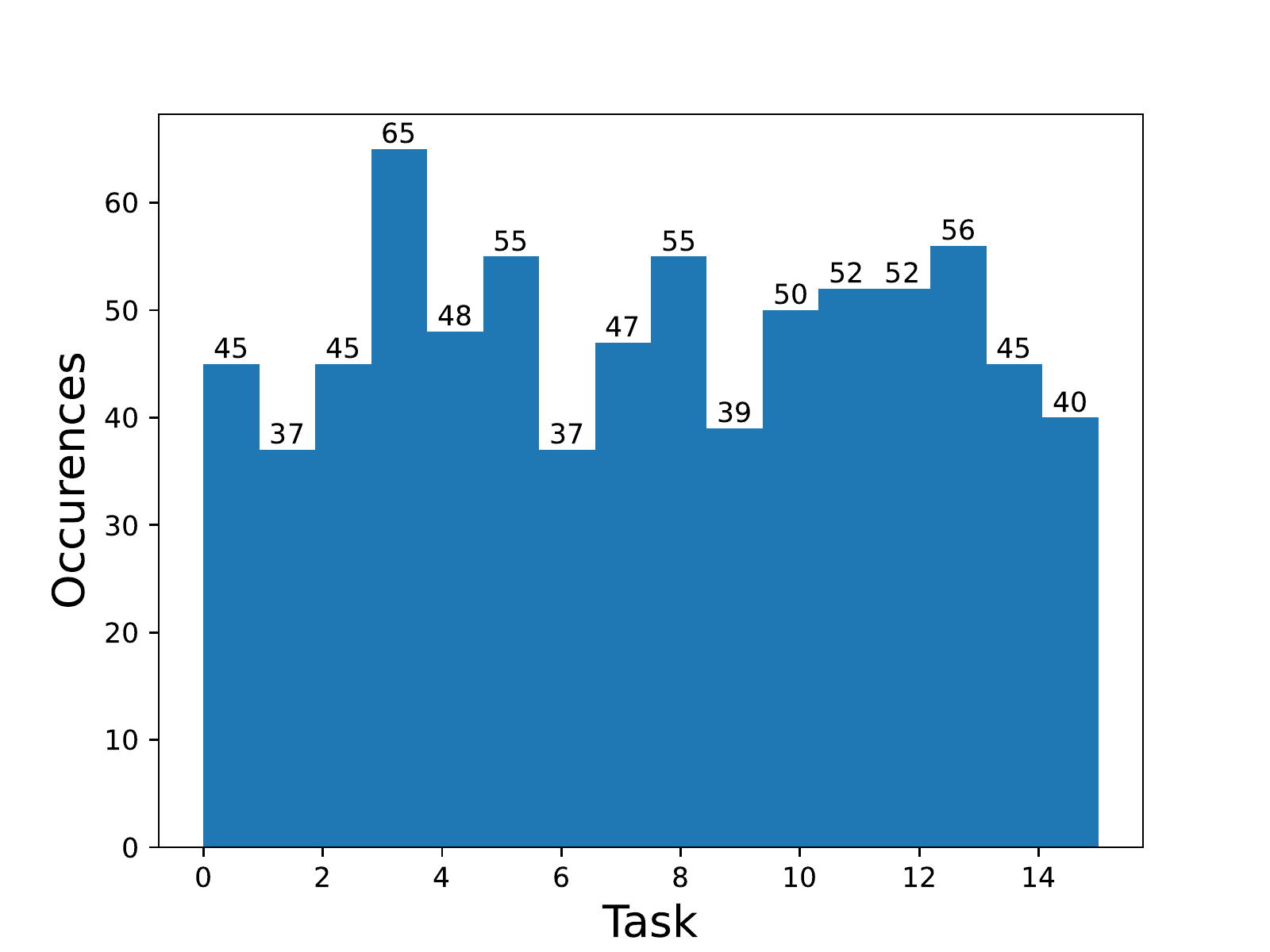} \\
    (B) \\
    \begin{tabular}{ccc}
        \includegraphics[width=0.4\textwidth]{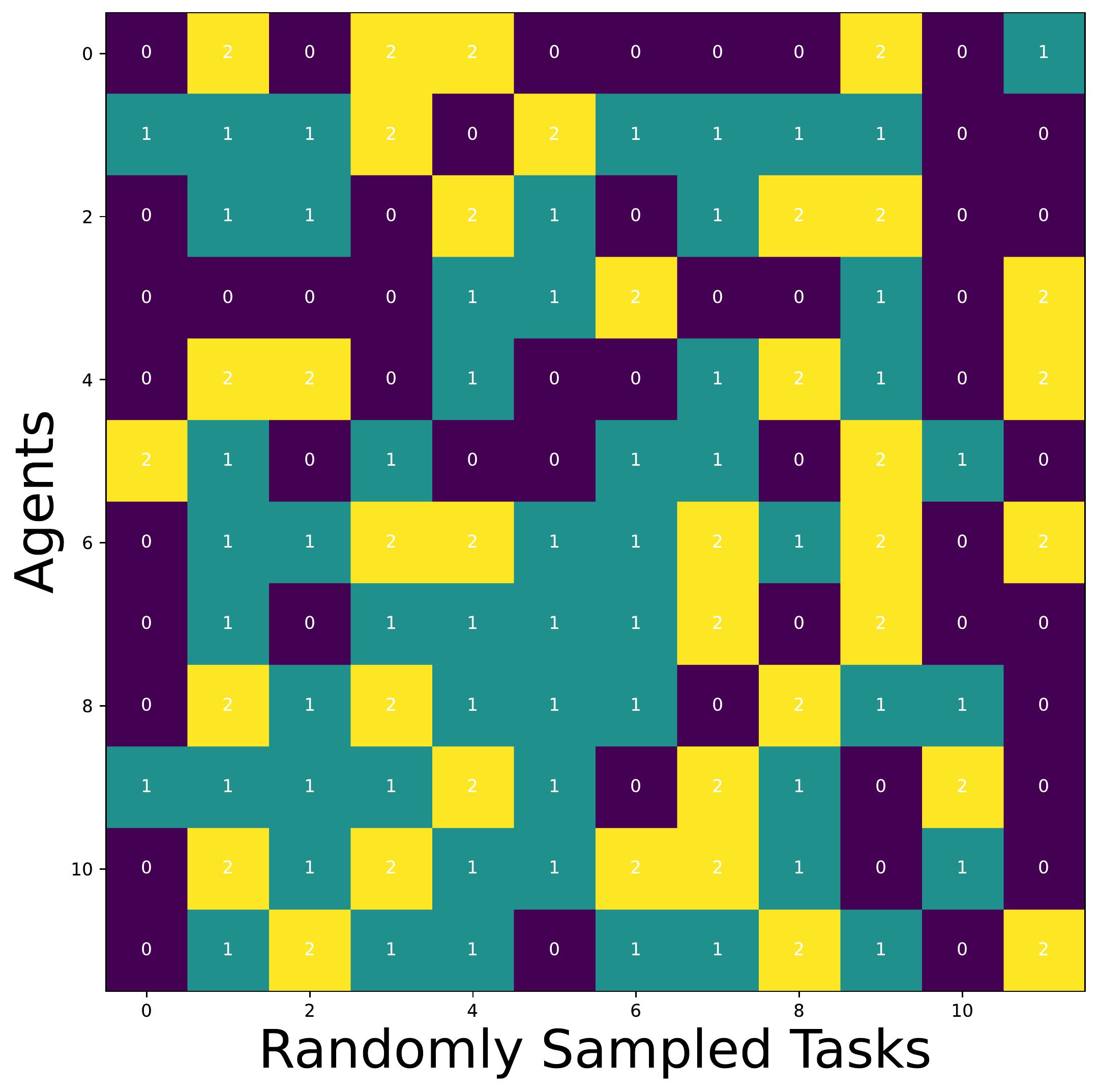} & 
        \includegraphics[width=0.56
        \textwidth]{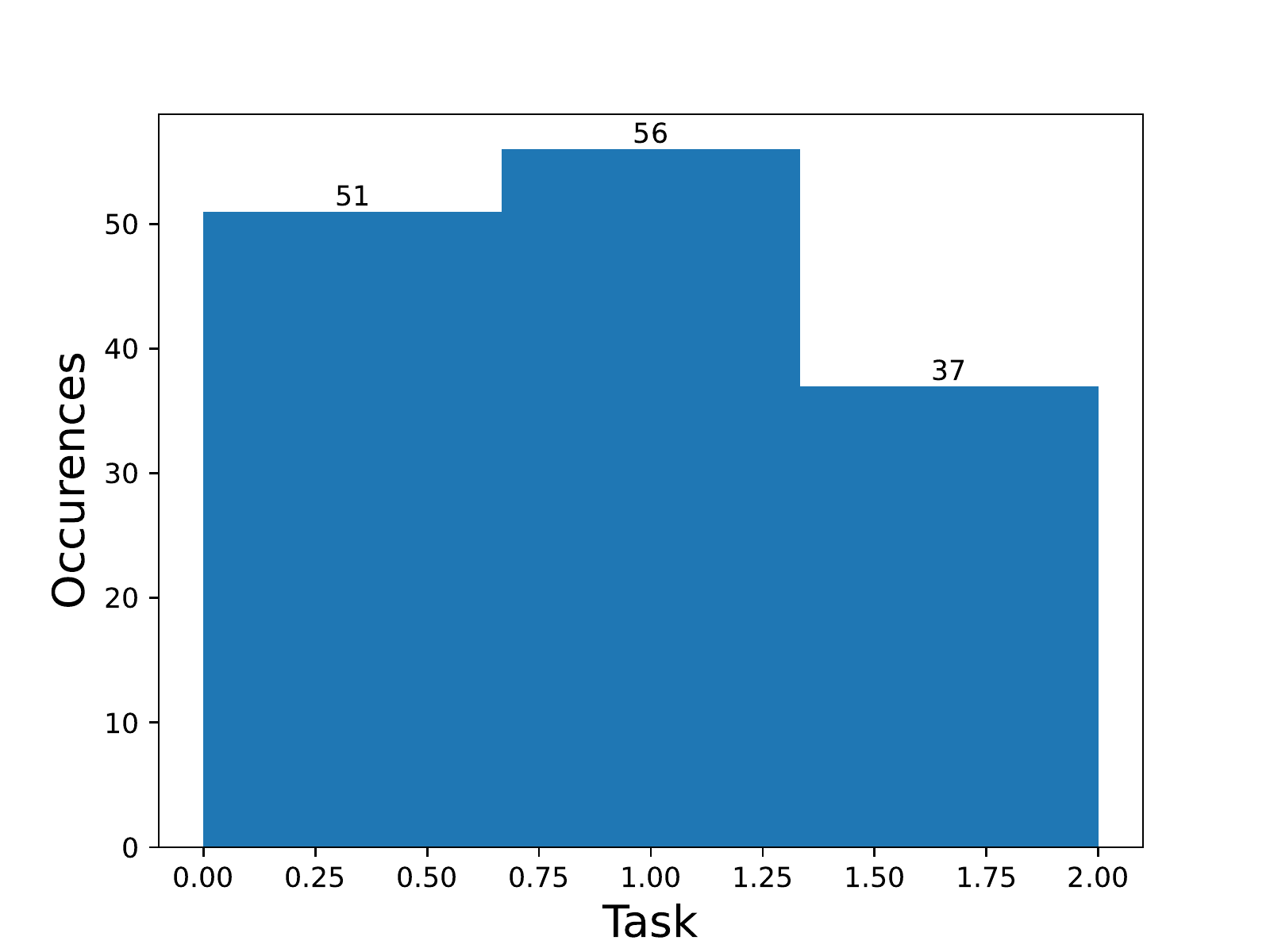}\\
        (C)&(D)\\
        
    \end{tabular}
    \caption{Distribution of randomly sampled tasks. (A) 12 sequences of tasks (one for each agent, plotted as 12 rows) are visualized with each task having a different color. The horizontal dimension is the time, showing here that there are 64 task changes. This task curriculum consists of tasks randomly sampled from a 16-task CT-graph. (B) The number of times each task is encountered across all learning time and agents. (C) 12 sequences of tasks are visualized for 12 task changes for 3 Minigrid task variations. (D) The number of times each task is encountered across all learning time and agents. For experiments shown in Fig.\ \ref{fig:commdrop} we use a modified curriculum in which the first agent uses a sequential repeating curriculum i.e., 0~15 for 4 cycles. The first agent also acts as the target of the local evaluation agent featured in figure \ref{fig:commdrop} and so it is a reflection of the learning dynamics of the first agent when it is connected to the rest of the collective with varying amounts of connection drops.}
    \label{fig:curriculum}
\end{figure}

In a typical execution with the CT-graph, tasks are sampled randomly. Fig.\ \ref{fig:curriculum} shows a typical curriculum that was visualized and analyzed. 

\subsection{Minigrid}
\label{apndx:env-minigrid}
\begin{figure}
    \centering
    \begin{subfigure}{0.25\textwidth}
        \centering
        \includegraphics[width=\textwidth]{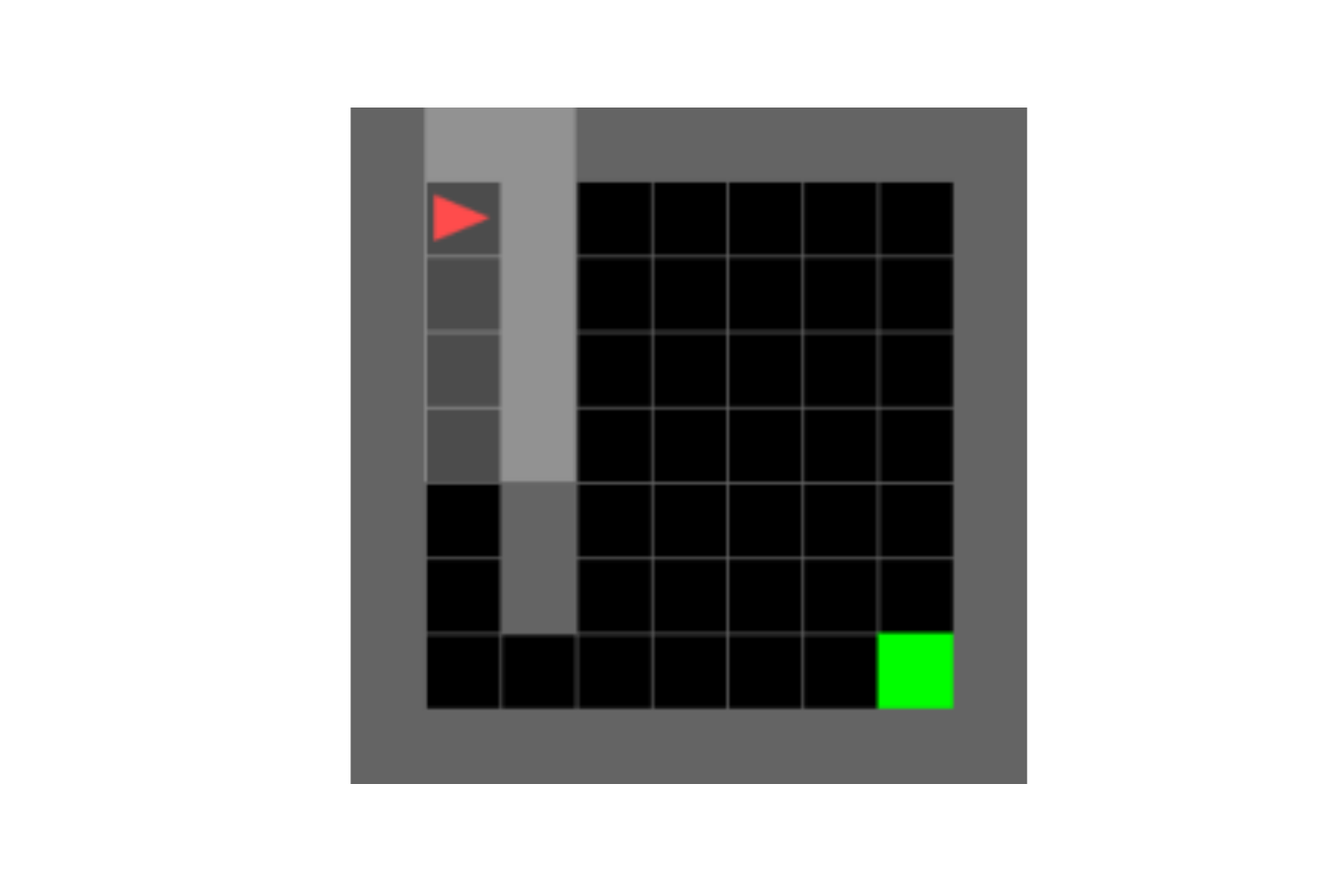}
    \end{subfigure}
    \begin{subfigure}{0.25\textwidth}
        \centering
        \includegraphics[width=\textwidth]{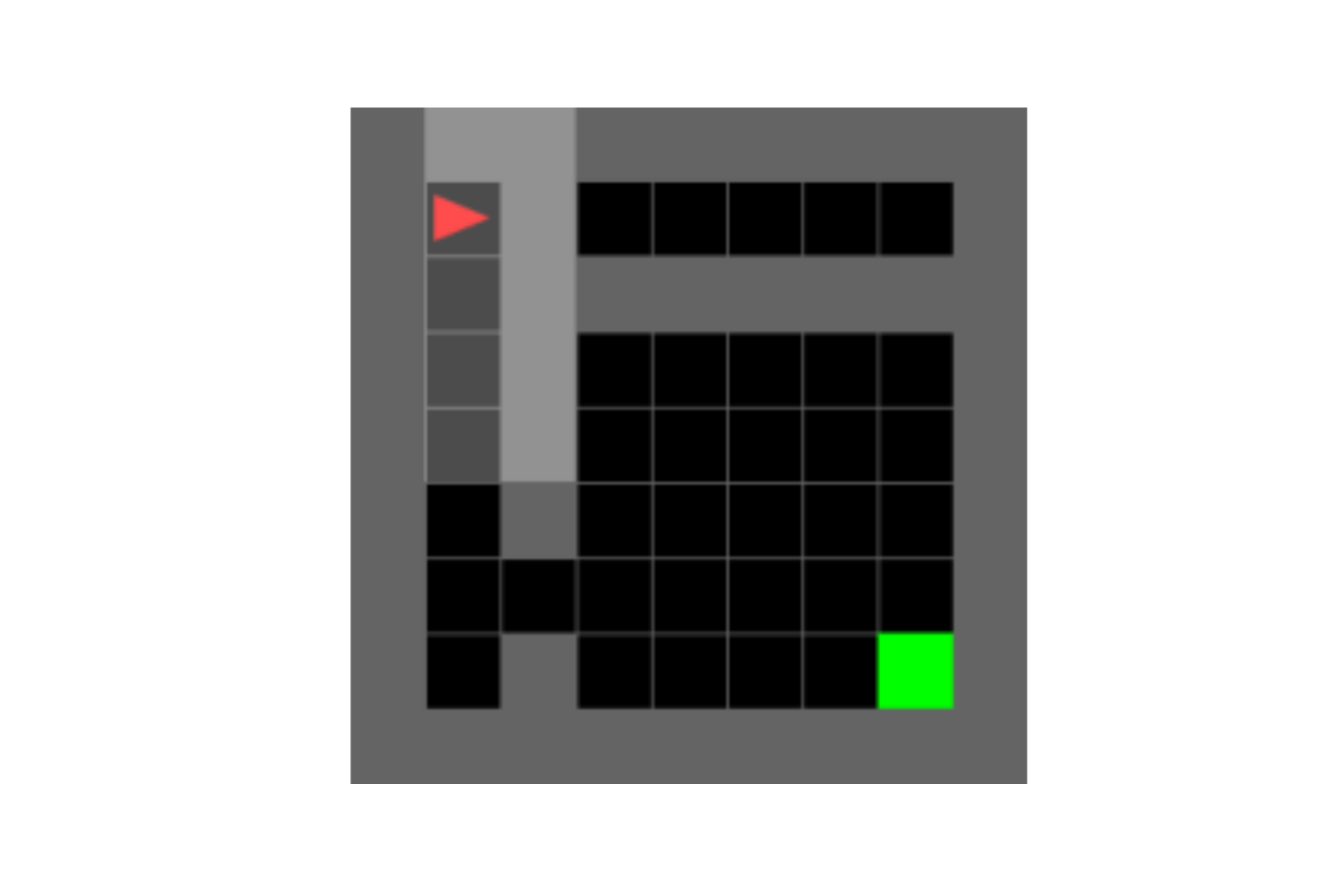}
    \end{subfigure}
    \begin{subfigure}{0.25\textwidth}
        \centering
        \includegraphics[width=\textwidth]{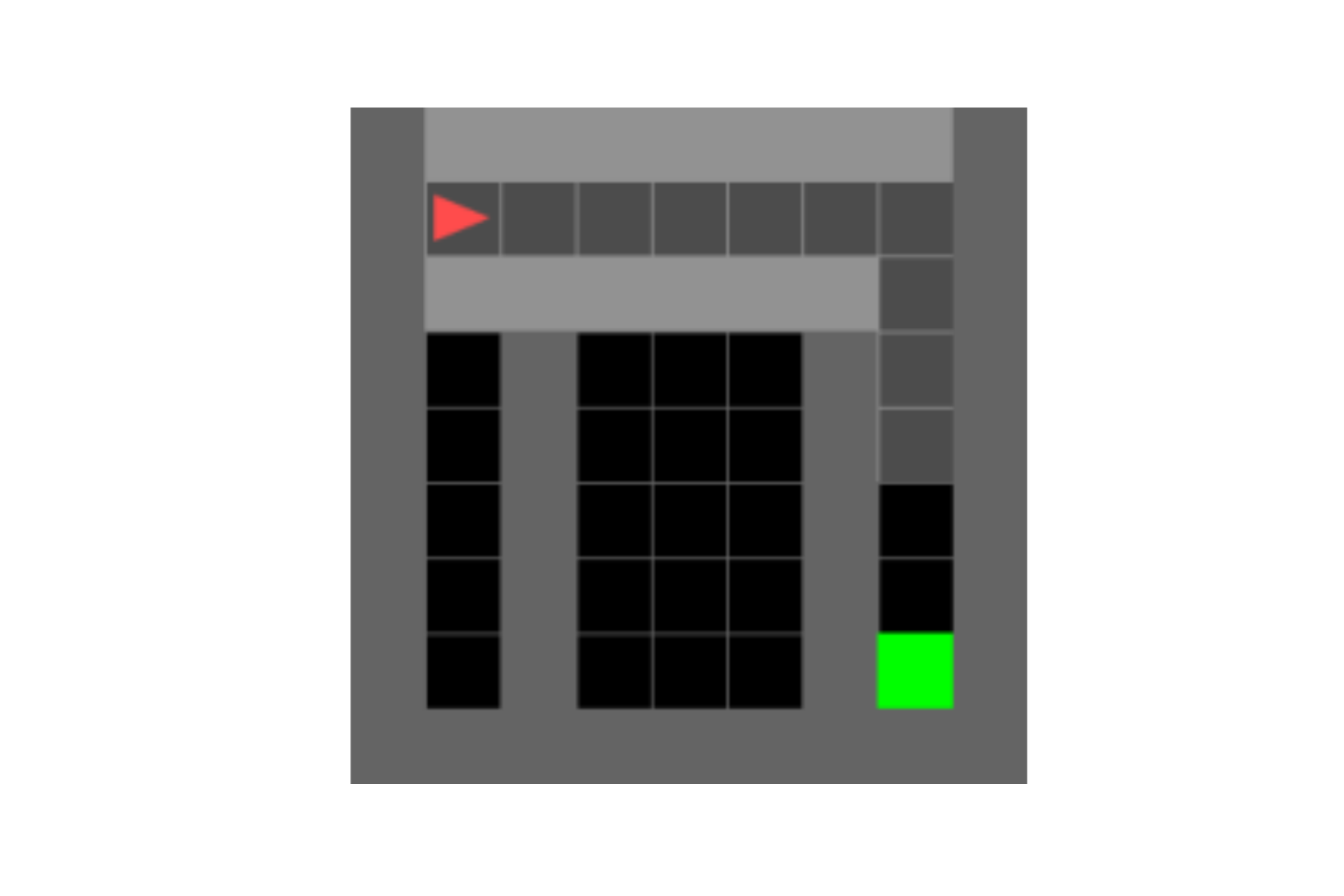}
    \end{subfigure}
 
    \caption{Visual representation of the 3 tasks in the \emph{MG3} curriculum. From left to right, one variant of each class: $\mathrm{SimpleCrossingS9N1}$, $\mathrm{SimpleCrossingS9N2}$, $\mathrm{SimpleCrossingS9N3}$. The agent (red) must navigate the environment with impassable walls (gray) and reach the goal state (green). The agent can only observe a limited portion of the environment, indicated by the highlighted squares in the images.}
    \label{fig:env-mingrid}
\end{figure}

The Minigrid environment \citep{gym_minigrid} is a grid world where an agent is required to navigate to the goal location. There are pre-defined grid worlds with sub-variants according to the random seed. A tensor of shape $7 \times 7 \times 3$ is used as input. The reward is defined as
\begin{equation}
    \label{eqn:minigrid-reward-fn}
    goal\_reward = 1 - 0.9 \times \frac{es}{ms}\quad,
\end{equation}
where $es$ is the number of steps taken to navigate to the goal and $ms$ is the maximum number of steps in an episode. The environment variations employed are: $\mathrm{SimpleCrossingS9N1}$, $\mathrm{SimpleCrossingS9N2}$, $\mathrm{SimpleCrossingS9N3}$. Fig.\ \ref{fig:env-mingrid} shows the five grid worlds used. 

\begin{figure}[t]
    \centering
    \begin{tabular}{ccc}
        \includegraphics[width=0.3\textwidth]{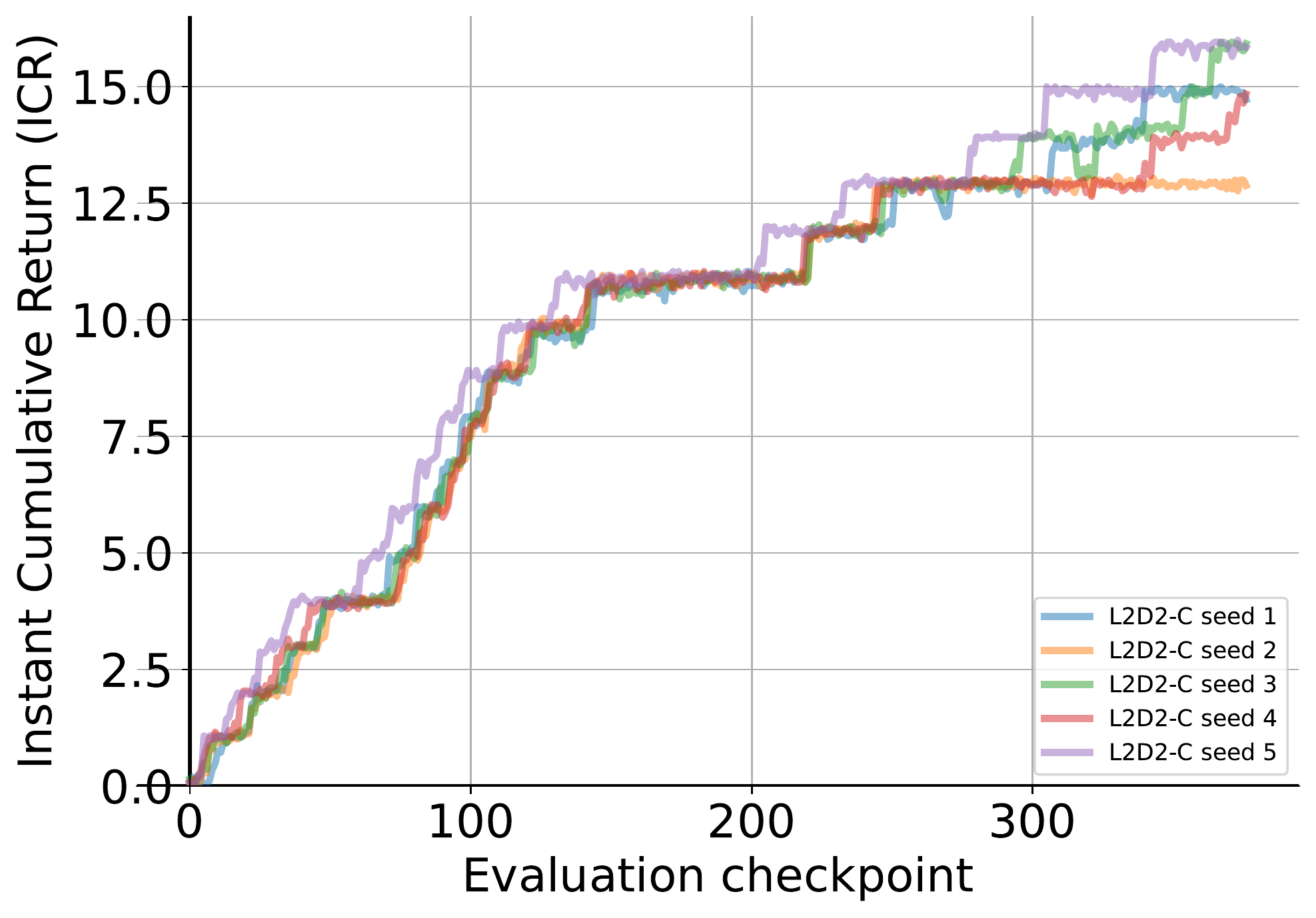} & 
        \includegraphics[width=0.3\textwidth]{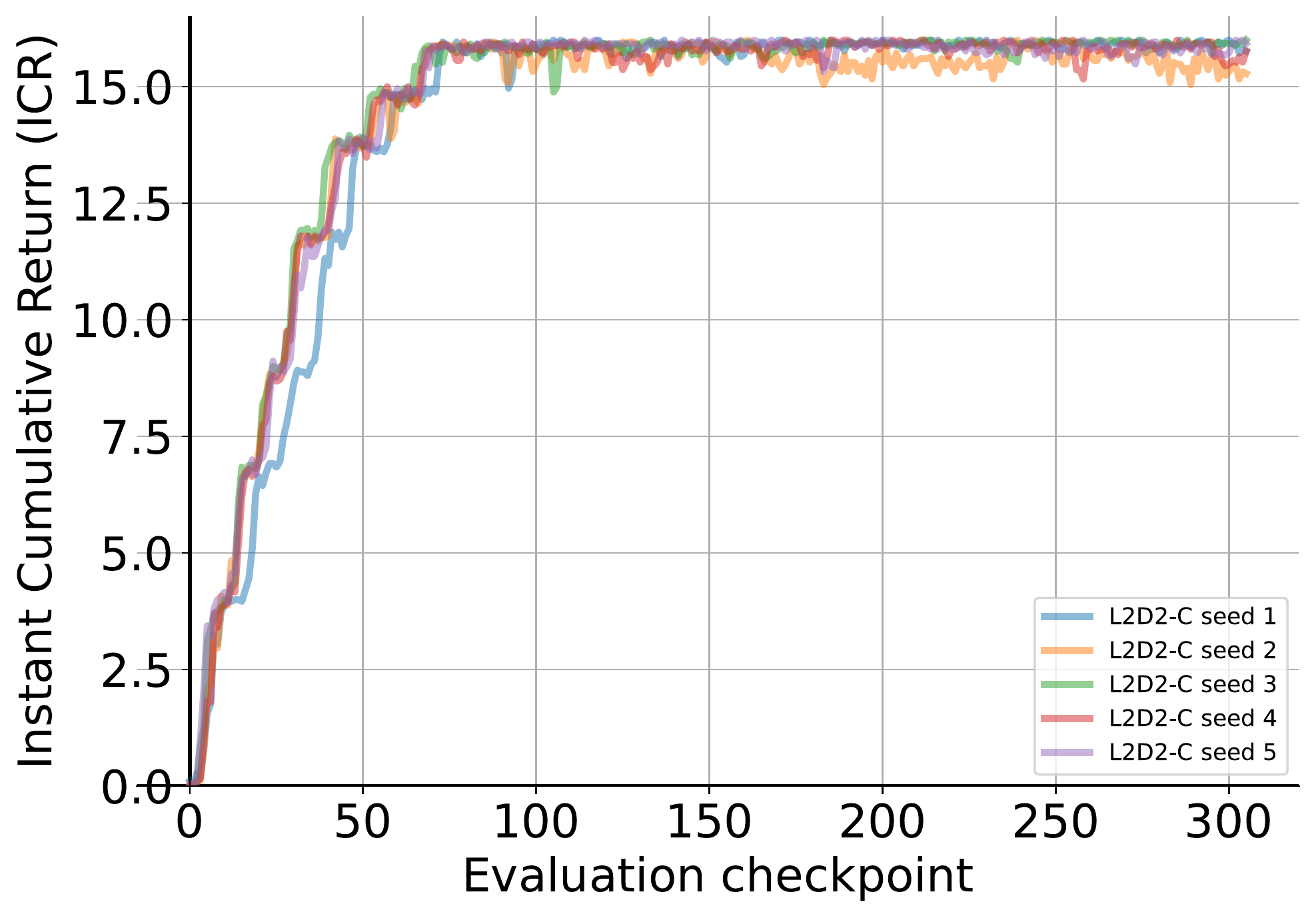} &
        \includegraphics[width=0.3\textwidth]{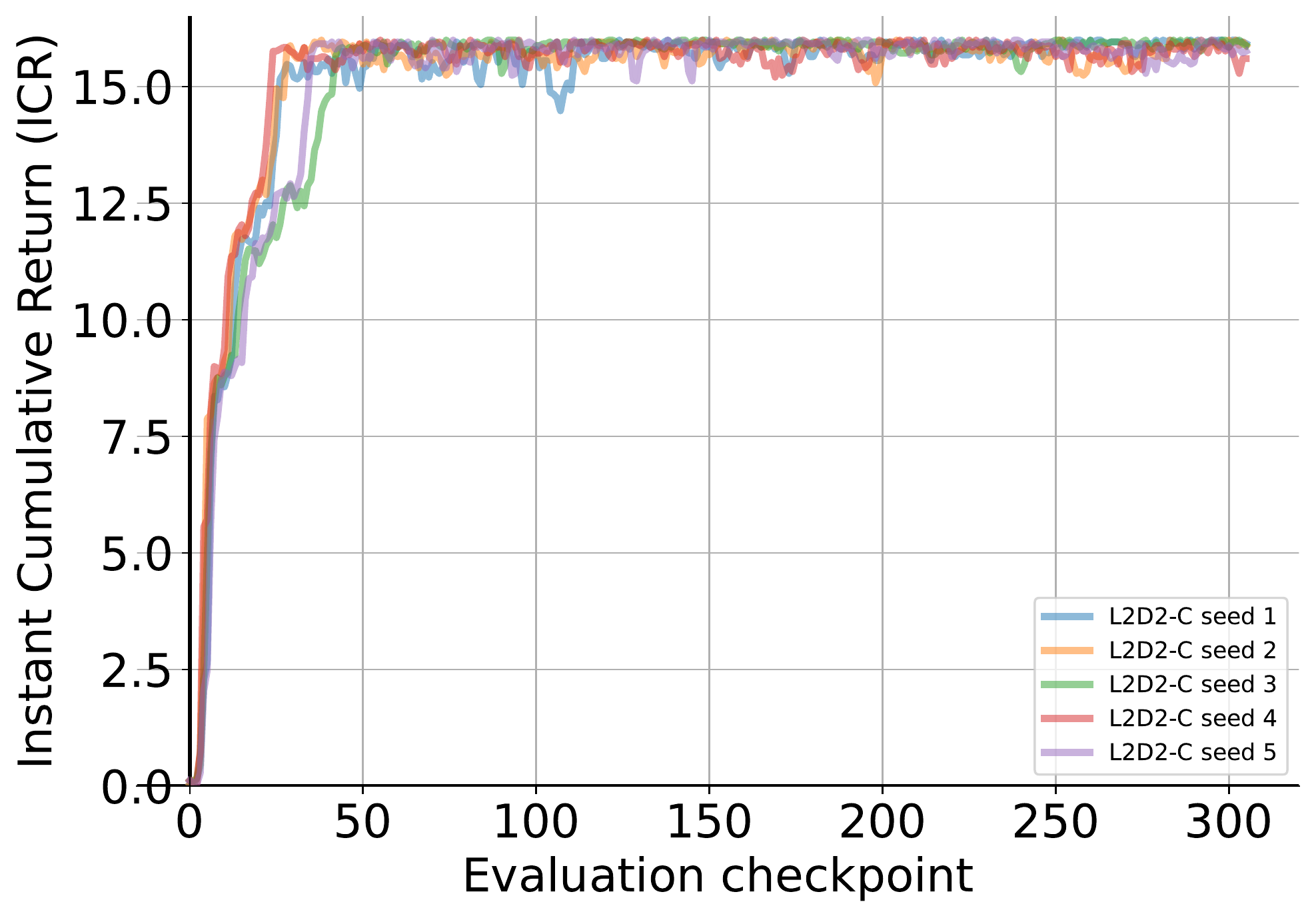}\\ 
        (A)&(B)&(C)\\
        \includegraphics[width=0.3\textwidth]{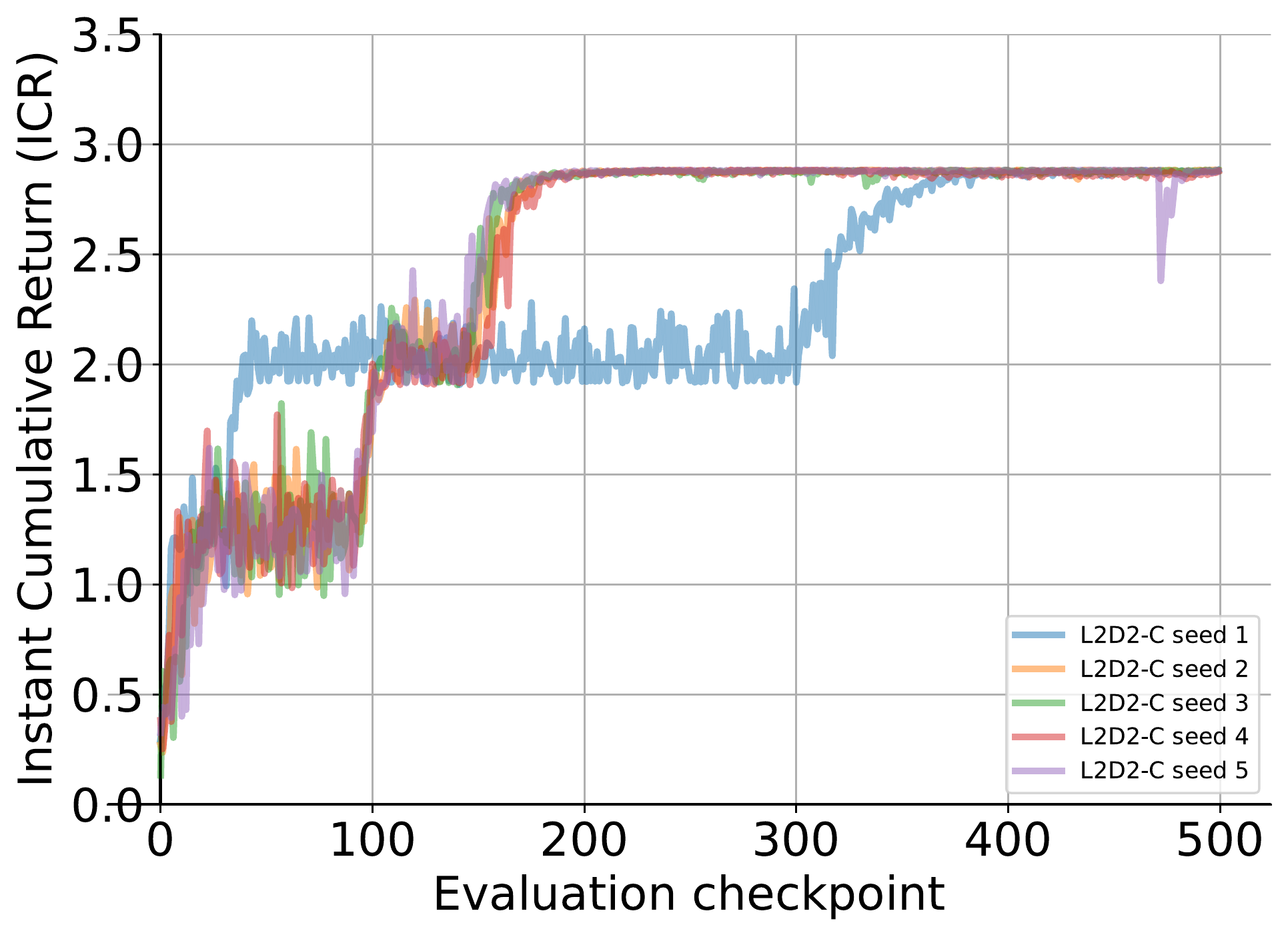} & \includegraphics[width=0.3\textwidth]{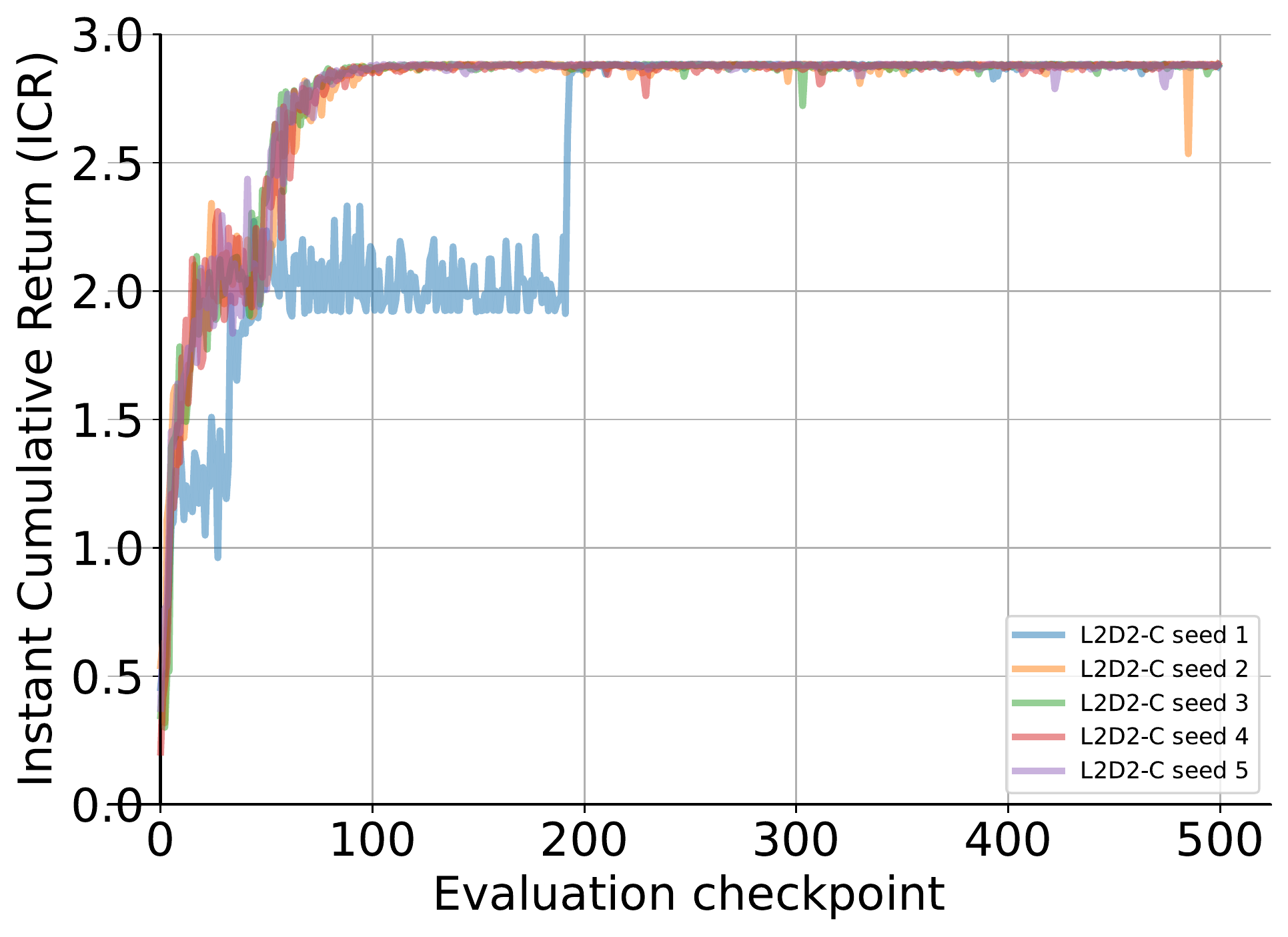} &
        \includegraphics[width=0.3\textwidth]{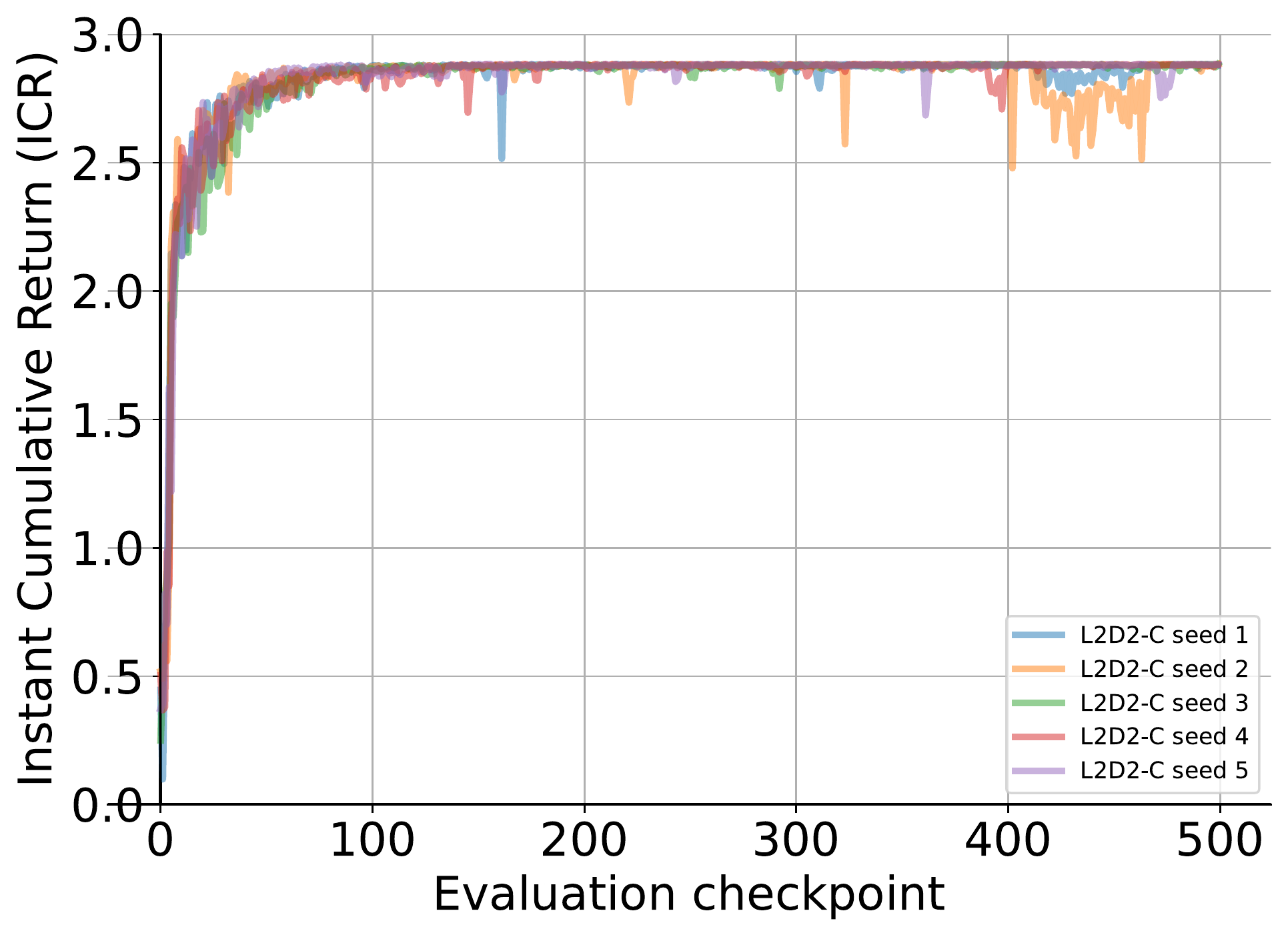}\\
        (D)&(E)&(F)\\
    \end{tabular}
    \caption{Individual seed runs of L2D2-C on CT-graph and Minigrid from Fig.\ \ref{fig:L2D2-C_comparison}. A total of 5 seeds were used for each experiment. The graphs show configurations of 1, 4, and 12 agents tested on the CT-graph environment (A, B, C) and similarly on the Minigrid environment (D, E, F).}
    \label{fig:seed_runs}
\end{figure}

\end{document}

%% file: math_commands.tex
\usepackage{amsmath,amsfonts,bm}

\def\eqref#1{equation~\ref{#1}}
\def\1{\bm{1}}

\DeclareMathAlphabet{\mathsfit}{\encodingdefault}{\sfdefault}{m}{sl}
\SetMathAlphabet{\mathsfit}{bold}{\encodingdefault}{\sfdefault}{bx}{n}